%% file: main.tex
\documentclass{article}
\usepackage[preprint]{neurips_2026}
\usepackage{xcolor}

\usepackage[utf8]{inputenc}
\usepackage[T1]{fontenc}
\usepackage{amsmath,amssymb,amsfonts}
\usepackage{booktabs}
\usepackage{multirow}
\usepackage{graphicx}
\usepackage{subcaption}
\usepackage{algorithm}
\usepackage{algorithmic}
\usepackage{hyperref}
\usepackage{cleveref}
\usepackage{xspace}
\usepackage{pifont}
\usepackage{microtype}
\usepackage{wrapfig}

\newcommand{\cmark}{\ding{51}}
\newcommand{\xmark}{\ding{55}}

\usepackage{amsthm}

\input{math_commands}

\title{TOUR: A Trajectory-Level Unlearning Benchmark for Offline Reinforcement Learning}

\author{%
  \textbf{Chaofan Pan}$^{1}$ \quad
  \textbf{Lingfei Ren}$^{1}$ \quad
  \textbf{Xiangyu Jiang}$^{1}$ \quad
  \textbf{Yanhua Li}$^{1}$ \quad
  \textbf{Xuemei Cao}$^{1}$ \\[0.5ex]
  \textbf{Xiangkun Wang}$^{1}$ \quad
  \textbf{Hao Yu}$^{1}$ \quad
  \textbf{Wei Wei}$^{2}$ \quad
  \textbf{Xin Yang}$^{1}$\thanks{Corresponding author.} \\[1ex]
  $^1$Southwestern University of Finance and Economics \\
  $^2$Shanxi University \\[0.5ex]
  \texttt{pan.chaofan@foxmail.com}\\
  \texttt{yangxin@swufe.edu.cn} \\
  \texttt{weiwei@sxu.edu.cn}
}

\begin{document}

\maketitle

\begin{abstract}
	\input{sections/0_abstract}
\end{abstract}

\input{sections/1_introduction}
\input{sections/2_related_work}
\input{sections/3_benchmark}
\input{sections/4_experiments}

\input{sections/5_discussion}
\input{sections/6_conclusion}

\bibliographystyle{unsrt}
\bibliography{references}

\newpage
\appendix
\input{sections/A_appendix}

\newpage

\end{document}

%% file: sections/0_abstract.tex
Offline Reinforcement Learning (RL) agents are trained on fixed behavioral trajectories, which makes trajectory-level deletion important when selected data must be removed after training. Evaluating such deletion is difficult because a lower membership score can reflect trajectory removal, residual memorization visible to another attack, or policy collapse that destroys useful behavior. We introduce \textbf{T}rajectory-level mem\textbf{O}rization and \textbf{U}nlearning in offline \textbf{R}L (\textbf{TOUR}), a benchmark that combines trajectory-level partitioning, matched non-member controls, retraining references, retained-performance anchors, and multi-attack privacy auditing. Across D4RL locomotion experiments and an exploratory AntMaze extension, TOUR shows that common deletion baselines have environment-dependent privacy-utility behavior. Retraining and fine-tuning often provide stronger retained-utility references than uniform GA+Refit, while TrajDeleter remains a useful comparator but is not uniformly stronger under the same audit. Reference-model, threshold, deviation, equivalence, action-error, representation-based, and query-limited attacks further show that a single likelihood-based membership score can overstate deletion quality. In the evaluated settings, conclusions about offline RL unlearning are therefore not stable under single-score auditing. 
They depend on matched non-member construction, retraining-relative calibration, attack family, retained utility, and explicit scope for diagnostic architecture or component-level evidence.

%% file: sections/1_introduction.tex
\section{Introduction}
\label{sec:intro}
\vspace{-0.2cm}

Offline Reinforcement Learning (RL) agents are increasingly used in domains such as autonomous driving and healthcare~\citep{prudencio2023survey,chen2024opportunities}, where policies are trained on fixed collections of sensitive behavioral trajectories~\citep{tian2023learning,jia2024offline}. Privacy regulations such as the General Data Protection Regulation and the California Consumer Privacy Act establish a ``right to be forgotten''~\citep{rigaki2023survey,liu2022right,wang2024comprehensive}, which creates pressure to remove the influence of selected trajectories after training. Machine unlearning provides tools for approximate data removal in supervised learning~\citep{bourtoule2021machine,guo2020certified,nguyen2025survey}, but offline RL adds a distinct evaluation challenge. Deletion should reduce trajectory-level evidence while preserving policy behavior on the retained task distribution.

This evaluation problem is more subtle than checking whether one membership score moves toward random guessing. A deletion update can reduce a forget-set score because it removes trajectory-specific information, but the same movement can also arise from policy collapse, degraded retained behavior, or residual evidence that is visible under a different attack. Retain-side diagnostics are also needed because an update can shift retained trajectories relative to matched non-members. These failure modes motivate the central question of this paper: \emph{How should trajectory-level unlearning in offline RL be evaluated so that apparent membership reduction is not confused with residual memorization, retain-side artifacts, or utility collapse?}

We answer this question with \textbf{T}rajectory-level mem\textbf{O}rization and \textbf{U}nlearning in offline \textbf{R}L (\textbf{TOUR}). TOUR is a benchmark centered on three MuJoCo continuous-control environments from D4RL~\citep{fu2020d4rl}, three locomotion data variants, and an exploratory AntMaze extension. It evaluates Decision Transformer (DT) policies~\citep{chen2021decision,liu2023constrained,wang2024critic}, MLP and Long Short-Term Memory (LSTM) policy families, and an Implicit Q-Learning (IQL)~\citep{kostrikov2022offline} companion analysis. The benchmark standardizes trajectory-level deletion, matched non-member construction, retraining references, retained-performance anchors, and a multi-family membership audit. The main comparisons use settings in which retained task performance makes privacy-utility interpretation meaningful.

\Cref{fig:hero} summarizes the evaluation protocol. TOUR first partitions trajectories into forget, retain, and matched non-member sets. It evaluates each update through forget-side membership, retain-side diagnostics, retained utility, and complementary attacks. The resulting evidence shows that common baselines fail in different ways. Uniform GA+Refit can move membership scores while losing utility, naive fine-tuning can appear private under the primary likelihood score without providing a complete audit on its own, and the TrajDeleter~\citep{liu2025trajdeleter} baseline has environment-dependent behavior with clearer retraining-relative residual signals in selected Walker2D settings.
The architecture and component-level experiments, together with supplementary controls reported in the appendix, provide diagnostic evidence. Their role is to reveal where the benchmark detects sensitivity rather than to prove a universal mechanism for selective deletion. In summary, our contributions are as follows:

\begin{figure}[t]
	\centering
	\includegraphics[width=0.95\textwidth]{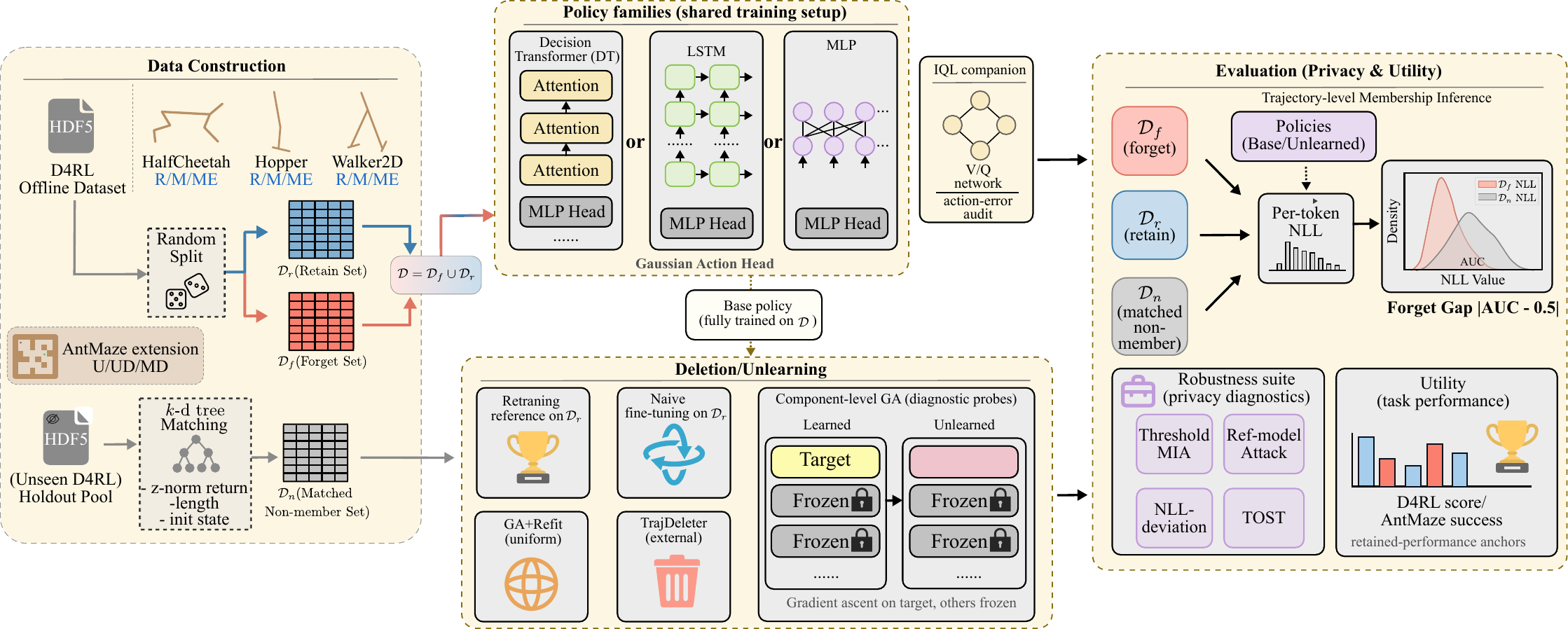}
	\caption{Overview of TOUR. The benchmark partitions trajectories into forget, retain, and matched non-member sets, applies deletion baselines to trained offline RL policies, and audits the updated policies through matched membership tests, retained-performance anchors, and complementary diagnostics. The protocol is designed to distinguish trajectory deletion evidence from policy collapse and single-attack artifacts.}
	\label{fig:hero}
	\vspace{-0.5cm}
\end{figure}

\begin{itemize}
	\item We introduce TOUR, a benchmark for trajectory-level memorization and unlearning in offline RL that combines matched non-member controls, retraining references, retained-performance anchors, and reproducible deletion baselines.
	\item We propose a multi-attack privacy-utility audit within TOUR in which deletion evidence requires reduced forget-set membership signal without retained-performance collapse or retain-side diagnostic failure.
	\item We provide benchmark findings showing that common deletion baselines can look private under a single score while failing reference-model or utility checks. Additional diagnostic experiments characterize architecture-associated and component-level sensitivity under bounded empirical scope.
\end{itemize}

%% file: sections/2_related_work.tex
\vspace{-0.2cm}
\section{Related Work}
\label{sec:related}
\vspace{-0.3cm}

\paragraph{Machine Unlearning.}
Machine unlearning seeks to remove the influence of selected training data without requiring full retraining~\citep{nguyen2025survey,liu2025rethinking}. Representative approaches include partition-based methods such as SISA~\citep{bourtoule2021machine}, certified removal for convex models~\citep{guo2020certified}, Fisher-information updates for deep networks~\citep{golatkar2020eternal}, and data bias-based plugins for continual learning~\citep{cao2025error}. A recurring challenge is that apparent deletion does not necessarily imply actual deletion, because gradient-based procedures can leave detectable traces of the removed data~\citep{thudi2022unrolling}. Recent work in vision and language modeling studies heterogeneous memorization by targeting attention components~\citep{pham2025lethevit}, pruning highly attributed modules~\citep{lin2025mape}, or isolating vulnerable spectral subspaces~\citep{fan2025sift}. TOUR differs in its primary goal. It uses structural diagnostics when useful, but the main contribution is an evaluation protocol for offline RL trajectory deletion.

\paragraph{Membership Inference and Privacy Auditing.}
Membership Inference Attacks (MIAs)~\citep{shokri2017membership} audit whether training data remain detectable after model training or unlearning. Threshold attacks can be effective~\citep{yeom2018privacy}, but a single attack family often gives an incomplete privacy assessment~\citep{salem2019mlleaks,song2021systematic}. For sequence models, example-level loss is a strong membership signal~\citep{carlini2022membership}. Offline RL introduces additional confounds because returns, trajectory lengths, and initial states can differ between deleted trajectories and non-members. TOUR therefore combines matched non-member sets with likelihood, reference-model, threshold, deviation, equivalence, action-error, and others to evaluate deletion quality as a privacy-utility profile rather than as a single score.

\paragraph{Offline RL and Model Architectures.}
Offline RL learns policies from fixed datasets without further environment interaction~\citep{levine2020offline}. Common policy families include autoregressive DTs~\citep{chen2021decision}, value-based methods such as Conservative Q-Learning (CQL)~\citep{kumar2020conservative} and IQL~\citep{kostrikov2022offline}, recurrent models such as LSTMs~\citep{hochreiter1997long}, and feedforward MLPs. Standard benchmarks such as D4RL~\citep{fu2020d4rl} focus on returns, whereas privacy and retained-data diagnostics under controlled deletion remain less studied.

\paragraph{RL-Specific Unlearning.}
Machine unlearning in RL remains limited because data are sequential and policy utility must be preserved after deletion. Prior work has studied removal of entire environments in online RL~\citep{ye2025reinforcement} and trajectory-level deletion through TrajDeleter~\citep{liu2025trajdeleter}. These studies focus primarily on deletion algorithms and utility retention. TOUR is complementary. It includes TrajDeleter as a baseline and asks how offline RL trajectory deletion should be audited without conflating membership reduction, residual memorization, retained-data artifacts, and utility collapse.


%% file: sections/3_benchmark.tex
\vspace{-0.2cm}
\section{Benchmark}
\label{sec:benchmark}
\vspace{-0.3cm}

\Cref{fig:hero} illustrates the overview of TOUR. The benchmark standardizes trajectory partitioning, controls major confounds in membership inference, and evaluates deletion quality through a structured privacy-utility evidence profile.

\subsection{Problem Formulation}
\label{sec:problem}
\vspace{-0.3cm}

Let $\Dall$ denote an offline dataset of trajectories collected by a behavior policy, where each trajectory $\tau = (s_0, a_0, r_0, \ldots, s_T, a_T, r_T)$ contains the sequence of states, actions, and rewards from a single episode. For DT-style conditioning, each timestep also carries a return-to-go variable $R_t = \sum_{t'=t}^{T} r_{t'}$. The dataset $\Dall$ is partitioned into a \emph{forget set} $\Df$ and a \emph{retain set} $\Dr$, with $\Dall = \Df \cup \Dr$.

Given a base policy $\pi_\theta$ trained on $\Dall$, the unlearning objective is to produce an updated policy $\pithetap$ that balances privacy and utility. From a privacy perspective, an attacker should not distinguish trajectories in $\Df$ from unseen non-members more reliably than random guessing. From a utility perspective, $\pithetap$ should retain effective behavior on the task distribution represented by $\Dr$.

TOUR treats privacy and utility as joint evidence. A credible evidence profile should show reduced forget-set membership evidence relative to the base policy while preserving retained-task performance and avoiding retain-side diagnostic failures. Apparent privacy improvements that arise primarily from policy collapse are reported as failure modes rather than as deletion evidence.

\vspace{-0.2cm}
\subsection{Environments and Data Construction}
\label{sec:envs}
\vspace{-0.3cm}

TOUR includes three D4RL continuous-control environments from the MuJoCo locomotion suite, HalfCheetah, Hopper, and Walker2D~\citep{fu2020d4rl}. Each environment is evaluated on the \texttt{medium-replay-v2} (R), \texttt{medium-v2} (M), and \texttt{medium-expert-v2} (ME) variants. We refer to the \texttt{medium-replay-v2} settings as replay variants. Utility in these environments is reported as the standard D4RL normalized score, where higher values indicate stronger task performance under the D4RL normalization. The main comparative analyses use locomotion settings in which the learned policies retain meaningful task performance. Replay and lower-power slices serve as stress tests or supplementary diagnostics when retained utility or precision is weaker. We also include a three-setting AntMaze extension comprising \texttt{antmaze-umaze-v2} (U), \texttt{antmaze-umaze-diverse-v2} (UD), and \texttt{antmaze-medium-diverse-v2} (MD). This extension is an exploratory navigation stress test. Its success-rate-derived utility is not directly comparable to locomotion returns, and the resulting evidence is not used to establish broader protocol validity beyond locomotion.

Unlike environment-level forgetting~\citep{ye2025reinforcement}, TOUR enforces strict trajectory-level partitioning: every episode of $\Dall$ belongs either to $\Df$ or to $\Dr$. By default, $\Df$ contains $10\%$ of the trajectories, sampled uniformly at random. Because membership can be confounded by trajectory returns, lengths, and initial states, a matched non-member set $\Dn$ is constructed from a disjoint holdout pool of D4RL datasets using $k$-d tree retrieval over $z$-normalized trajectory features. These features consist of episode return, trajectory length, and the initial-state coordinates used in the matching diagnostics of Appendix~\ref{sec:appendix_matching_power}. This procedure maintains $|\Dn| = |\Df|$ while reducing the largest observable trajectory-level differences, but it does not eliminate all residual confounds. Appendix~\ref{sec:appendix_matching_power} therefore reports both balance diagnostics and precision summaries.

\vspace{-0.2cm}
\subsection{Policy Families and Model Architectures}
\label{sec:models}
\vspace{-0.3cm}

To investigate the relationship between architectural bias and memorization, we evaluate three policy families under a shared training budget of $100$K gradient steps and a uniform context length $K=20$ timesteps. These are a 3-layer DT with a Generative Pre-trained Transformer 2 (GPT-2)-style causal self-attention backbone~\citep{radford2019language}, a 3-layer LSTM, and an MLP baseline without explicit sequence aggregation. The DT and LSTM provide autoregressive or recurrent sequence modeling, whereas the MLP serves as a feedforward comparison. To avoid ambiguity, the term MLP baseline refers only to this standalone policy architecture. We use Gaussian action heads that predict diagonal Gaussian policies over mean and log-variance outputs, enabling exact per-token NLL computation during privacy auditing. During unlearning, variance parameters remain frozen to avoid variance inflation and to keep the likelihood audit comparable across updates. An IQL comparison~\citep{kostrikov2022offline} provides a value-based reference evaluated under a trajectory-level action-error audit because IQL does not parameterize the Gaussian action distribution needed for exact per-token NLL evaluation.

\vspace{-0.2cm}
\subsection{Multi-Attack Privacy Auditing Protocol}
\label{sec:tmi}
\vspace{-0.3cm}

Because relying on a single metric can be misleading~\citep{song2021systematic}, TOUR evaluates unlearning through a likelihood-centered Trajectory-level Membership Inference (TMI) protocol.

\paragraph{Primary Attack.}
For a given trajectory $\tau$, the mean per-timestep NLL is
\begin{equation}
	\NLL(\tau; \theta) = \frac{1}{T+1} \sum_{t=0}^{T} -\log \pi_\theta(a_t \mid s_t, R_t, t),
	\label{eq:nll}
\end{equation}
where $a_t$, $s_t$, and $R_t$ denote the action, state, and return-to-go at timestep $t$, and the final argument of $\pi_\theta$ is the timestep index used by DT-style conditioning. Trained models typically assign lower NLL values to seen trajectories, so $-\mathrm{NLL}$ serves as the membership score. We report the Area Under the Receiver Operating Characteristic Curve (AUC-ROC, hereafter AUC) between $\Df$ and $\Dn$. Privacy is summarized by the \textbf{forget gap}, defined as $|\AUCf - 0.5|$, where $\AUCf$ denotes the forget-versus-negative AUC and smaller values indicate weaker membership evidence. Values above $0.5$ mean that the model fits forget trajectories more confidently than matched non-members, whereas values below $0.5$ invert that ordering. We use the direction-agnostic gap because either direction still indicates separability once the attack score is allowed to flip sign. We also report $\AUCr$, the retain-versus-negative AUC, as a diagnostic validity check.

\paragraph{Robustness Suite.}
To reduce overfitting to the primary metric, TOUR includes four complementary likelihood-based checks: a threshold MIA~\citep{yeom2018privacy}, a reference-model attack against the retraining reference, an NLL-deviation AUC (the absolute deviation of each trajectory's NLL from the mean retain-set NLL), and Two One-Sided Tests (TOST) equivalence testing with a margin of $\varepsilon = 0.1$. In the reference-model attack, the retrained model serves as a calibrated reference and the membership score is $\NLL(\tau;\theta_{\mathrm{ref}}) - \NLL(\tau;\theta')$, so higher values indicate trajectories that the target model fits more confidently than the retraining reference. The retraining reference is an operational floor rather than a certified zero-influence oracle. It is not expected to achieve exactly $0.5$ because the forget set and matched non-member set can retain residual distributional differences after matching. The TOST analysis tests whether the forget-set AUC is statistically equivalent to random guessing, namely $0.5$, within the stated margin. Since these checks still focus mainly on trajectory likelihoods, action-error, representation-based, and query-limited shadow-model attacks are additionally reported in Appendices~\ref{sec:appendix_action_error}, \ref{sec:appendix_representation_attack}, and \ref{sec:appendix_shadow_query_attack}. These attacks serve as exploratory diagnostics depending on the attack family and data regime rather than as direct replacements.

\vspace{-0.2cm}
\subsection{Benchmark Baselines and Diagnostic Probes}
\label{sec:blocks}
\vspace{-0.3cm}

TOUR compares five policy conditions: a base model $\pi_\theta$ trained on $\Dall$, a retraining reference policy $\piretrain$ trained from scratch on $\Dr$, a naive Fine-Tuned (FT) model continued on $\Dr$, a uniform Gradient-Ascent update followed by Refitting (GA+Refit), and a DT-adapted TrajDeleter baseline. Retraining and naive fine-tuning inherit the default training configuration: batch size $64$, learning rate $10^{-4}$, weight decay $10^{-4}$, linear warmup for $10$K steps, and a $100$K-step budget. In the ascent stage of GA+Refit, the updated policy $\pi_{\theta'}$ minimizes
\begin{equation}
	\mathcal{L}_{\mathrm{GA}}(\theta') = -\E_{\tau \sim \Df}\!\left[\NLL(\tau; \theta')\right] + \lambda \, \mathbb{E}_{\tau \sim \Dr}\!\left[\KL\big(\pi_{\theta}(\cdot\mid\tau)\,\|\,\pi_{\theta'}(\cdot\mid\tau)\big)\right],
	\label{eq:ga_refit_loss}
\end{equation}
where $\lambda$ is the ascent-strength hyperparameter that controls the tradeoff between deletion pressure on $\Df$ and behavioral preservation on $\Dr$. Unless stated otherwise, the benchmark uses $\lambda=1.0$, ascent learning rate $10^{-4}$, gradient clipping at $0.25$, and a default $500$-step ascent budget. Then it refits a reinitialized action head on $\Dr$ for $10$K steps at learning rate $10^{-4}$ using batch size $64$.
The TrajDeleter baseline is treated as an external structured deletion comparator \citep{liu2025trajdeleter} rather than as a proposed method.
Its configuration uses $\alpha=1.0$, $\beta=2.0$, $100$ steps of the first stage, and $1,000$ steps of the second stage, where $\alpha$ weights the forget-directed loss in the first stage and $\beta$ weights the retain-side anchor KL term in the second stage.

For component-level analysis, utility is assessed through two complementary criteria. The first criterion is a relative utility budget against the matched uniform baseline: a target is feasible only when it does not underperform the corresponding uniform update at the same ascent budget. The second criterion reports retained performance ratios against both the retraining reference and the original base policy. Appendix~\ref{sec:appendix_selective_criteria} gives the formal definitions.

To diagnose whether deletion pressure has different effects across network regions, TOUR includes \emph{component-level GA} as a diagnostic or exploratory probe. Instead of updating the entire model, this procedure freezes most parameters and applies gradient ascent only to a targeted component. A random-matched control with an identical parameter count to the target layer helps distinguish component-associated effects from generic small-update effects. These comparisons are interpreted jointly with the explicit retained-performance criteria and are not used as standalone evidence for a deployable component-level unlearning method.

\vspace{-0.2cm}
\subsection{Benchmark Evidence Structure}
\vspace{-0.2cm}
Our benchmark separates three levels of evidence.
The required evidence consists of the canonical forget gap under the DT likelihood audit, the retain-versus-negative diagnostic AUC, and explicit retained-performance anchors. Calibrated diagnostics, including the reference-model gap, threshold attack, deviation attack, and TOST, refine the interpretation of residual memorization within the same audit family. Exploratory evidence includes low-power settings, the AntMaze extension, the IQL companion analysis, and broad component-level sweeps. This separation keeps the main claim tied to a structured evidence profile while exposing structural patterns that warrant further study.

%% file: sections/4_experiments.tex
\vspace{-0.2cm}
\section{Experimental Results}
\label{sec:experiments}
\vspace{-0.3cm}

\input{figures/TABLE_1_benchmark}

TOUR tests whether trajectory deletion can be evaluated without conflating membership reduction, retained-data artifacts, and utility collapse. The results show that privacy conclusions depend jointly on matched controls, attack family, and retained-performance anchors. Unless stated otherwise, result tables report mean aggregates over three seeds.

\subsection{Benchmark Validity and Deletion Baselines}
\label{sec:bench_results}
\vspace{-0.3cm}

The evaluation begins with the locomotion baselines in \Cref{tab:benchmark}. The primary privacy quantity is the direction-agnostic forget gap.
The 95\% intervals summarize seed-level uncertainty for the aggregate method score and should not be read as independent-trajectory uncertainty, while Appendix~\ref{sec:appendix_matching_power} reports matched-pair counts, interval widths, detectable-gap diagnostics, and the quartile-level evidence in \Cref{fig:quartile}.

Across most locomotion settings, the base DT exhibits a detectable membership signal. In the main locomotion comparisons, retraining references and naive fine-tuning often move the forget-set AUC close to random guessing while preserving more utility than uniform GA+Refit, whereas the DT-adapted TrajDeleter baseline remains environment-dependent under the same audit. TrajDeleter broadens the comparison through an external structured deletion procedure, but its canonical configuration does not yield a uniformly stronger privacy-utility profile than the retraining-based references. Replay variants broaden coverage, but the main comparative claims come from settings in which retained task performance and statistical precision keep the tradeoff interpretable.
Uniform GA+Refit often reduces utility sharply and does not provide a reliable privacy advantage. In several settings, including HalfCheetah (M) and Hopper (M), the forget-set AUC falls below $0.5$. This value indicates an inverted ordering rather than erasure because the forget set remains separable from matched non-members after score reversal. The benchmark therefore treats privacy improvements obtained mainly through policy collapse as failures.

Appendix~\ref{sec:appendix_antmaze_extension} reports the AntMaze extension as supplementary navigation evidence under a separate success-rate-derived utility scale. The results distinguish a stronger U-Maze variant from weaker U-Maze-Diverse and Medium-Diverse variants and therefore provide a stress test of the audit outside the locomotion suite without implying score equivalence to the locomotion benchmark.

\subsection{Single Membership Scores Are Insufficient}
\label{sec:multi_attack}
\vspace{-0.3cm}

\input{figures/TABLE_multi_attack}

\Cref{tab:multi_attack} applies the likelihood-centered audit suite to test robustness. The table focuses on selected high-utility locomotion settings that combine nontrivial base membership signal, retained-performance viability, and sufficient matched-pair precision. Appendix~\ref{sec:appendix_multi_attack} extends this view with complementary attack families rather than with a complete NLL matrix. This restricted threat model extends the primary per-token NLL AUC with threshold, reference-model, deviation, and equivalence checks while remaining within the same likelihood family. Naive FT shows that an NLL score close to the random-guessing region is not by itself a complete audit, although its retraining-relative gaps are small in this selected slice. The clearest reference-model failures instead occur for TrajDeleter in Walker2D (ME), where near-random primary NLL scores can coexist with large retraining-relative residual signals. A single metric can therefore overstate deletion quality.

The other baselines support the same broad interpretation. The base DT exhibits membership signals across attacks, retraining provides the operational privacy reference, and uniform GA+Refit still leaves residual signals or loses utility depending on the environment. The DT-adapted TrajDeleter baseline is included in \Cref{tab:multi_attack} as an external structured deletion comparator. Its residual signals under reference-model calibration show that procedure-aware deletion still requires retraining-relative and retained-utility checks.

Appendices~\ref{sec:appendix_action_error}, \ref{sec:appendix_representation_attack}, and \ref{sec:appendix_shadow_query_attack} extend this analysis with action-error, representation-based, and query-limited shadow-model attacks. These diagnostics do not overturn the main likelihood audit, but they show that method rankings remain attack-family-dependent and that stronger gradient-ascent pressure does not produce a consistent privacy advantage across environments and audit families.

\subsection{Architecture Results as Descriptive Diagnostics}
\label{sec:arch}
\vspace{-0.3cm}

To investigate architecture-associated differences, we compare DT, MLP, and LSTM policies on the replay variants. As summarized in \Cref{tab:architecture_summary}, all rows come from a unified backbone comparison configuration with the same trajectory split, matched non-member construction, context length, and likelihood-based TMI audit. The architecture comparison is included to expose how audit conclusions change under policy-family changes, not to isolate architecture as a causal factor. The MLP and LSTM baselines both remain below the DT on the reported forget-gap diagnostic, but their relative ordering and separation vary substantially across environments. Appendix~\ref{sec:appendix_non_dt_baselines} extends the scope with an IQL analysis based on an action-error audit rather than the DT likelihood score. Hopper and Walker2D provide the main comparative settings, while HalfCheetah (R) is retained as a lower-power reference. Appendices~\ref{sec:appendix_architecture_control} and \ref{sec:appendix_non_dt_baselines} add narrower controls and supplementary non-DT evidence. Taken together, the results support an architecture-associated difference under the evaluated training configurations, but not a capacity-matched causal claim, a utility-matched comparison, or a score-equivalent DT-versus-IQL ranking.

\input{figures/TABLE_architecture_summary}

\subsection{Component-Level Updates as Diagnostic Probes}
\label{sec:selective}
\vspace{-0.3cm}

\begin{wrapfigure}[18]{r}{0.5\textwidth}
	\centering
	\includegraphics[width=0.5\textwidth]{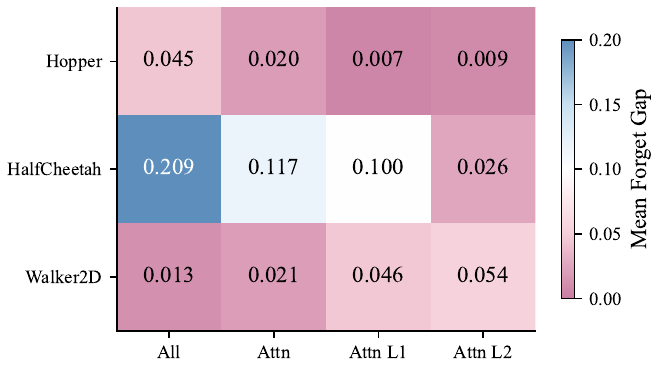}
	\caption{Mean forget gap by environment and component-level GA target. Component-level updates expose environment-dependent sensitivity, while smaller relative gaps do not necessarily imply deletion with retained performance.}
	\label{fig:heatmap}
\end{wrapfigure}

Component-level GA is used as a diagnostic probe for update sensitivity inside DT policies. The probe is evaluated on all attention layers jointly and on individual attention layers. \Cref{fig:heatmap} summarizes the resulting forget gaps, while Appendices~\ref{sec:appendix_tost} and \ref{sec:appendix_selective_budget} provide stability checks and utility-budget comparisons. Because TrajDeleter changes the deletion procedure rather than isolating individual DT components, it is treated as a benchmark baseline in \Cref{sec:bench_results} rather than as part of the component-level sweep.

The component-level sweep reveals substantial heterogeneity across environments. In the replay variants analysis, some attention-layer targets reduce the forget gap relative to matched uniform updates under a relative utility budget, but these gains are not sufficient under explicit retained-performance criteria (\Cref{tab:layer_cv,tab:oracle,tab:selective_budget}). The retained-utility check yields a narrower signal outside this replay slice. Hopper provides the clearest replay-setting diagnostic case: Attention Layer 1 at $250$ steps reaches a forget-set AUC of $0.499$ with D4RL score $11.8$ in the full sweep (\Cref{tab:full_selective}). This value is close to random guessing only under severe utility loss, so it should be read as a policy-collapse warning rather than as a successful deletion result. The corresponding Walker2D and HalfCheetah settings remain either farther from random guessing or deep in a low-utility regime. The evidence therefore supports component sensitivity rather than a general selective-unlearning method.

\Cref{tab:oracle} summarizes the replay-setting component-level policy under the zero additional utility loss slice alongside the matched uniform baseline and retraining reference. Appendix~\ref{sec:appendix_selective_criteria} and \Cref{fig:pareto} report the broader privacy-utility frontier and utility-budget view from the full sweep. The absolute utility ratios show that the relative utility criterion is weaker than the explicit retained-performance criteria. Component-level GA is therefore better viewed as a diagnostic probe than as a practical unlearning method, and the benchmark contract does not count relative-budget feasibility alone as sufficient deletion evidence.
Appendices~\ref{sec:appendix_random}, \ref{sec:appendix_localization}, and \ref{sec:appendix_fisher_baseline} provide additional random-mask, gradient, and Fisher controls. These checks sharpen the interpretation of the component-level sweep, but they do not alter the main conclusion of this section.

\input{figures/TABLE_2_oracle}

%% file: figures/TABLE_1_benchmark.tex
\begin{table}[htp]
\centering
\renewcommand{\arraystretch}{0.8}
\caption{Benchmark results across R/M/ME variants of three locomotion environments. The main comparative claims rely on settings with meaningful retained task performance. Utility denotes the D4RL normalized score. Forget-Set AUC is retained as an auxiliary direction-sensitive quantity. Retain AUC is a diagnostic attack score, not a utility metric. Pair counts and detectable-gap diagnostics are reported separately in \Cref{tab:matching_power}.}
\label{tab:benchmark}
\resizebox{0.9\textwidth}{!}{%
\begin{tabular}{llrrrcr}
\toprule
Environment & Method & Utility $\uparrow$ & Forget Gap $\downarrow$ & Forget AUC & 95\% CI & Retain AUC \\
\midrule
\multirow{5}{*}{Walker2D (ME)} & Base DT & 93.11 & 0.015 & 0.515 & [0.513, 0.518] & 0.538 \\
 & Retrain Ref. & 87.11 & 0.015 & 0.485 & [0.484, 0.487] & 0.545 \\
 & Naive FT & 85.46 & 0.017 & 0.483 & [0.482, 0.483] & 0.534 \\
 & GA+Refit & 0.00 & 0.015 & 0.515 & [0.513, 0.519] & 0.500 \\
 & TrajDeleter & 92.73 & \textbf{0.012} & 0.512 & [0.510, 0.515] & 0.536 \\
\midrule
\multirow{5}{*}{HalfCheetah (ME)} & Base DT & 64.13 & 0.140 & 0.640 & [0.635, 0.646] & 0.563 \\
 & Retrain Ref. & 59.82 & 0.058 & 0.558 & [0.553, 0.561] & 0.575 \\
 & Naive FT & 60.63 & \textbf{0.054} & 0.554 & [0.553, 0.555] & 0.552 \\
 & GA+Refit & 4.70 & 0.094 & 0.594 & [0.585, 0.604] & 0.496 \\
 & TrajDeleter & 58.74 & 0.126 & 0.626 & [0.621, 0.633] & 0.554 \\
\midrule
\multirow{5}{*}{Hopper (ME)} & Base DT & 44.04 & 0.049 & 0.549 & [0.548, 0.550] & 0.507 \\
 & Retrain Ref. & 47.70 & 0.004 & 0.504 & [0.499, 0.509] & 0.513 \\
 & Naive FT & 46.01 & \textbf{0.002} & 0.502 & [0.501, 0.503] & 0.499 \\
 & GA+Refit & 8.78 & 0.033 & 0.467 & [0.465, 0.468] & 0.489 \\
 & TrajDeleter & 42.63 & 0.046 & 0.546 & [0.543, 0.549] & 0.512 \\
\midrule
\midrule
\multirow{5}{*}{Walker2D (M)} & Base DT & 71.92 & 0.121 & 0.621 & [0.621, 0.622] & 0.597 \\
 & Retrain Ref. & 67.62 & \textbf{0.051} & 0.551 & [0.550, 0.553] & 0.607 \\
 & Naive FT & 77.63 & 0.053 & 0.553 & [0.552, 0.554] & 0.584 \\
 & GA+Refit & 13.50 & 0.107 & 0.607 & [0.603, 0.610] & 0.494 \\
 & TrajDeleter & 67.90 & 0.107 & 0.607 & [0.605, 0.609] & 0.583 \\
\midrule
\multirow{5}{*}{HalfCheetah (M)} & Base DT & 42.77 & 0.197 & 0.697 & [0.693, 0.705] & 0.724 \\
 & Retrain Ref. & 42.69 & 0.043 & 0.457 & [0.444, 0.465] & 0.753 \\
 & Naive FT & 42.95 & \textbf{0.034} & 0.466 & [0.451, 0.475] & 0.717 \\
 & GA+Refit & 4.73 & 0.274 & 0.226 & [0.197, 0.253] & 0.502 \\
 & TrajDeleter & 41.58 & 0.160 & 0.660 & [0.656, 0.661] & 0.694 \\
\midrule
\multirow{5}{*}{Hopper (M)} & Base DT & 43.20 & 0.053 & 0.553 & [0.550, 0.555] & 0.519 \\
 & Retrain Ref. & 45.55 & \textbf{0.016} & 0.484 & [0.480, 0.488] & 0.525 \\
 & Naive FT & 46.87 & 0.020 & 0.480 & [0.478, 0.484] & 0.511 \\
 & GA+Refit & 17.04 & 0.125 & 0.375 & [0.339, 0.400] & 0.550 \\
 & TrajDeleter & 47.15 & 0.046 & 0.546 & [0.542, 0.549] & 0.520 \\
\midrule
\midrule
\multirow{5}{*}{Walker2D (R)} & Base DT & 25.66 & 0.125 & 0.625 & [0.622, 0.630] & 0.528 \\
 & Retrain Ref. & 44.09 & \textbf{0.020} & 0.520 & [0.518, 0.522] & 0.506 \\
 & Naive FT & 36.47 & 0.028 & 0.528 & [0.523, 0.532] & 0.517 \\
 & GA+Refit & 10.50 & 0.089 & 0.589 & [0.586, 0.595] & 0.438 \\
 & TrajDeleter & 26.26 & 0.071 & 0.571 & [0.570, 0.572] & 0.493 \\
\midrule
\multirow{5}{*}{HalfCheetah (R)} & Base DT & 32.29 & 0.224 & 0.724 & [0.719, 0.730] & 0.695 \\
 & Retrain Ref. & 30.61 & 0.002 & 0.498 & [0.495, 0.505] & 0.718 \\
 & Naive FT & 32.14 & \textbf{0.000} & 0.500 & [0.490, 0.505] & 0.718 \\
 & GA+Refit & 7.96 & 0.284 & 0.784 & [0.776, 0.791] & 0.456 \\
 & TrajDeleter & 35.37 & 0.156 & 0.656 & [0.643, 0.668] & 0.684 \\
\midrule
\multirow{5}{*}{Hopper (R)} & Base DT & 9.26 & 0.014 & 0.514 & [0.496, 0.524] & 0.581 \\
 & Retrain Ref. & 25.17 & 0.018 & 0.518 & [0.514, 0.520] & 0.619 \\
 & Naive FT & 22.15 & 0.016 & 0.516 & [0.511, 0.523] & 0.605 \\
 & GA+Refit & 11.20 & 0.010 & 0.510 & [0.464, 0.592] & 0.510 \\
 & TrajDeleter & 17.37 & \textbf{0.009} & 0.509 & [0.503, 0.512] & 0.620 \\
\bottomrule
\end{tabular}%
}
\vspace{-0.5cm}
\end{table}

%% file: figures/TABLE_multi_attack.tex
\begin{table}[htp]
\centering
\renewcommand{\arraystretch}{0.8}
\caption{Multi-attack privacy audit on the selected high-utility comparative settings aggregated across seeds. Four membership inference attacks are applied to each method. NLL: per-timestep NLL AUC. Thr: threshold MIA balanced accuracy~\citep{yeom2018privacy}. Ref: reference-model calibrated AUC from the score $\NLL(\tau;\theta_{\mathrm{ref}})-\NLL(\tau;\theta')$. Dev: NLL-deviation AUC based on absolute deviation from the mean NLL of the retain set. Values closer to 0.5 indicate weaker membership evidence. The retraining-reference row reports ``---'' for Ref AUC because the retrained model is itself the reference model in that attack, so the calibrated score is undefined. TOST reports whether all available seeds pass the equivalence test at significance level 0.05. All$<\varepsilon$ reports whether every reported likelihood-family attack gap remains below $\varepsilon=0.1$ across all available seeds. These two columns need not agree because one is a statistical test and the other is a threshold condition. The NLL AUC values are recomputed inside the multi-attack pipeline and then aggregated across seeds, so they need not match the benchmark-table forget-set AUC summary exactly.}
\label{tab:multi_attack}
\resizebox{0.9\textwidth}{!}{%
\begin{tabular}{llccccccc}
\toprule
Environment & Method & Pairs & NLL AUC & Thr BA & Ref AUC & Dev AUC & TOST & All$<\varepsilon$ \\
\midrule
\multirow{5}{*}{HalfCheetah (ME)} & Base DT & 140 & 0.640 & 0.598 & 0.988 & 0.443 & \xmark & \xmark \\
 & Retrain Ref. & 140 & 0.558 & 0.550 & --- & 0.443 & \xmark & \cmark \\
 & Naive FT & 140 & \textbf{0.554} & 0.545 & \textbf{0.460} & \textbf{0.453} & \xmark & \cmark \\
 & GA+Refit & 140 & 0.594 & \textbf{0.531} & 0.607 & 0.578 & \xmark & \xmark \\
 & TrajDeleter & 140 & 0.626 & 0.598 & 0.961 & 0.440 & \xmark & \xmark \\
\midrule
\multirow{5}{*}{Walker2D (ME)} & Base DT & 153 & 0.515 & 0.505 & 0.861 & 0.477 & \cmark & \xmark \\
 & Retrain Ref. & 153 & 0.485 & 0.497 & --- & 0.495 & \cmark & \cmark \\
 & Naive FT & 153 & 0.483 & 0.490 & \textbf{0.490} & \textbf{0.497} & \cmark & \cmark \\
 & GA+Refit & 153 & 0.515 & 0.512 & 0.549 & 0.487 & \cmark & \cmark \\
 & TrajDeleter & 153 & \textbf{0.512} & \textbf{0.501} & 0.853 & 0.474 & \cmark & \xmark \\
\midrule
\multirow{5}{*}{Walker2D (M)} & Base DT & 83 & 0.621 & 0.598 & 0.999 & 0.459 & \xmark & \xmark \\
 & Retrain Ref. & 83 & \textbf{0.551} & \textbf{0.534} & --- & 0.478 & \xmark & \cmark \\
 & Naive FT & 83 & 0.553 & 0.542 & \textbf{0.470} & \textbf{0.479} & \xmark & \cmark \\
 & GA+Refit & 83 & 0.607 & 0.602 & 0.627 & 0.449 & \xmark & \xmark \\
 & TrajDeleter & 83 & 0.607 & 0.578 & 0.911 & 0.464 & \xmark & \xmark \\
\bottomrule
\end{tabular}%
}
\vspace{-0.5cm}
\end{table}

%% file: figures/TABLE_architecture_summary.tex
\begin{table}[htp]
\centering
\renewcommand{\arraystretch}{0.8}
\caption{Architecture diagnostic for the replay variants under a unified backbone comparison configuration. All rows use backbone-policy evaluations with the same trajectory split, matched controls, context length, and likelihood-based TMI audit. Parameter counts, utility, forget gap, and interval width are shown together so that architecture-associated differences are interpreted with the main confounds in view rather than as capacity-matched or utility-matched causal isolation. The interval width is computed from the displayed CI endpoints in \Cref{tab:architecture}.}
\label{tab:architecture_summary}
\resizebox{0.6\linewidth}{!}{%
\begin{tabular}{llrrrr}
\toprule
Environment & Model & Params & Utility $\uparrow$ & Forget Gap $\downarrow$ & CI Width \\
\midrule
\multirow{3}{*}{Hopper} & DT & 726K & 22.35 & 0.023 & 0.003 \\
 & MLP & 69K & 15.28 & \textbf{0.003} & 0.001 \\
 & LSTM & 593K & 1.70 & 0.015 & 0.004 \\
\midrule
\multirow{3}{*}{HalfCheetah} & DT & 727K & 36.77 & 0.228 & 0.016 \\
 & MLP & 71K & 35.32 & \textbf{0.034} & 0.005 \\
 & LSTM & 594K & 1.81 & 0.155 & 0.025 \\
\midrule
\multirow{3}{*}{Walker2D} & DT & 727K & 45.22 & 0.093 & 0.004 \\
 & MLP & 71K & 20.46 & 0.055 & 0.000 \\
 & LSTM & 594K & 0.31 & \textbf{0.018} & 0.001 \\
\bottomrule
\end{tabular}%
}
\vspace{-0.4cm}
\end{table}

%% file: figures/TABLE_2_oracle.tex
\begin{table}[htp]
\centering
\renewcommand{\arraystretch}{0.8}
\caption{Component-level GA under an explicit zero additional utility loss budget relative to the matched uniform baseline. The component-level results use the leave-one-seed-out cross-validation target, aggregated over step choices that satisfy $\Delta$utility $\leq 0$ relative to the matched uniform update. This aggregation differs from the per-held-out-split CV gaps in \Cref{tab:layer_cv}, which report one cross-validation split at a time. \textbf{Score/Retrain} and \textbf{Score/Base} report explicit retained-performance ratios relative to the retraining reference and the base DT, so that relative feasibility with respect to uniform updates is not conflated with a stronger retained-performance criterion.}
\label{tab:oracle}
\resizebox{1\textwidth}{!}{%
\begin{tabular}{llp{2.5cm}rrrr}
\toprule
Environment & Method & Relative-budget feasible & Forget Gap $\downarrow$ & Utility $\uparrow$ & Score/Retrain & Score/Base \\
\midrule
\multirow{3}{*}{Hopper} & Retraining reference & --- & 0.018 & 25.17 & 1.00 & 2.72 \\
 & \textbf{Selective (held-out CV)} & 9/15 & \textbf{0.006} & 15.09 & 0.60 & 1.63 \\
 & Uniform (matched baseline) & --- & 0.043 & 6.24 & 0.25 & 0.67 \\
\midrule
\multirow{3}{*}{HalfCheetah} & \textbf{Retraining reference} & --- & \textbf{0.002} & 30.61 & 1.00 & 0.95 \\
 & Selective (held-out CV) & 2/9 & 0.015 & 1.52 & 0.05 & 0.05 \\
 & Uniform (matched baseline) & --- & 0.186 & 1.26 & 0.04 & 0.04 \\
\midrule
\multirow{3}{*}{Walker2D} & Retraining reference & --- & 0.020 & 44.09 & 1.00 & 1.72 \\
 & Selective (held-out CV) & 0/9 & 0.028 & 9.43 & 0.21 & 0.37 \\
 & \textbf{Uniform (matched baseline)} & --- & \textbf{0.013} & 2.18 & 0.05 & 0.08 \\
\bottomrule
\end{tabular}%
}
\vspace{-0.5cm}
\end{table}

%% file: sections/5_discussion.tex
\vspace{-0.3cm}
\section{Discussion}
\label{sec:discussion}
\vspace{-0.3cm}

TOUR supports a benchmark claim: trajectory-level unlearning in offline RL should be evaluated through matched controls, retained-performance anchors, and multiple attack families. This protocol distinguishes cases in which forget-set membership evidence decreases together with retained utility from cases in which a near-random privacy score is explained by policy collapse or by attack-specific calibration artifacts. Its role is to make privacy-utility evidence comparable across deletion procedures rather than to certify a single best deletion algorithm.

The current evidence is also narrower than a mechanism paper would require. Architecture comparisons show that privacy behavior differs across model families under the evaluated shared training configuration, but utility, capacity, optimization, and audit family are not fully disentangled. The IQL comparison broadens the empirical scope, yet it uses trajectory-level action error rather than the likelihood-centered DT protocol. Component-level GA shows that update sensitivity varies across DT components and environments, but the retained-performance checks keep this evidence diagnostic rather than algorithmic. These results motivate more controlled follow-up studies, but they do not identify architecture as the unique cause of memorization or establish a deployable component-level unlearning procedure.

The unlearning results show why future methods should report evidence in tiers. Uniform GA+Refit often moves policies into a low-utility regime, as shown in \Cref{tab:benchmark}, while retraining and fine-tuning preserve more task performance in several settings. The multi-attack audit further shows that the primary NLL score alone can miss residual signals, especially under reference-model calibration. The TrajDeleter comparison points to the same conclusion from a different direction: an external structured deletion baseline still requires retraining-relative and retained-utility checks before a privacy claim is convincing under the evaluated audit.

The main limitations are the single-head DT architecture, the remaining confounds in cross-architecture comparisons, and the limited statistical power in lower-power settings such as HalfCheetah and Hopper replay. AntMaze is exploratory because it uses a separate success-rate utility scale, the navigation settings show strong environment-dependent difficulty, and the lower-power U-Maze-Diverse and Medium-Diverse cases still yield coarse privacy uncertainty. The attack suite also covers only a subset of practical attackers. Additional limitations are discussed in Appendix~\ref{sec:appendix_limitations}.

%% file: sections/6_conclusion.tex
\vspace{-0.2cm}
\section{Conclusion}
\label{sec:conclusion}
\vspace{-0.3cm}

This paper introduces TOUR, a benchmark for trajectory-level memorization and unlearning in offline RL. TOUR combines matched non-member controls, retraining references, retained-performance anchors, and complementary membership attacks so that deletion quality is evaluated as a privacy-utility evidence profile rather than as a single membership score.
Across the evaluated D4RL experiments, common deletion baselines fail in different ways. Some updates reduce apparent membership evidence only while damaging policy utility, whereas others look private under one likelihood score but leave residual signals under retraining-relative or complementary attacks. Architecture and component-level experiments provide descriptive diagnostics, but the current evidence supports setting-specific interpretation rather than a universal component-localization mechanism. Future offline RL unlearning methods should therefore report forget-set membership evidence together with retraining-relative residual signals, retain-side diagnostics, retained task performance, and attack-family scope of each deletion claim.

%% file: sections/A_appendix.tex
\section{Environment and Configuration}
\label{sec:appendix_env_config}

\subsection{Environment Overview}
\label{sec:appendix_env_overview}

TOUR spans three offline RL locomotion environments with distinct control and reward structures. HalfCheetah, Hopper, and Walker2D are continuous-control tasks from D4RL~\citep{fu2020d4rl}. HalfCheetah is a planar running task with smooth forward locomotion, Hopper requires stable single-leg hopping under repeated impacts, and Walker2D requires coordinated bipedal movement with stronger balance constraints. These three environments are evaluated on the \texttt{medium-replay-v2} (R), \texttt{medium-v2} (M), and \texttt{medium-expert-v2} (ME) variants to test whether privacy conclusions persist across shifts in data quality within the locomotion suite. The benchmark also includes an AntMaze extension across \texttt{antmaze-umaze-v2} (U), \texttt{antmaze-umaze-diverse-v2} (UD), and \texttt{antmaze-medium-diverse-v2} (MD). Unlike the locomotion tasks, these navigation settings are reported with a success-rate-derived utility scale. They are interpreted as an exploratory navigation stress test rather than as a direct continuation of the locomotion score scale or as a validation of broader protocol generality.

\subsection{Device and Runtime Details}
\label{sec:appendix_device_time}

All experiments were conducted on a single workstation equipped with an AMD Ryzen 9 7950X 16-Core Processor, 96\,GB of system memory, and a single NVIDIA GeForce RTX 4070 Ti SUPER graphics card with 16\,GB of video memory. The GPU driver version is 555.42.02 and the CUDA version is 12.4. The software stack comprises Python 3.12, PyTorch 2.2, Gymnasium with MuJoCo support version 1.2.3, scikit-learn 1.8.0, NumPy 2.4.3, and SciPy 1.14.0. All training and evaluation scripts run on the GPU by default; CPU is used only for data preprocessing and matching diagnostics.

Each base Decision Transformer training run requires 100\,K gradient steps with batch size 64, which takes approximately 15--25 minutes on the described hardware. Retraining references and naive fine-tuning follow the same training budget and thus share a comparable wall-clock cost. The GA+Refit procedure applies 500 ascent steps followed by 10\,K refit steps on the reinitialized action head, completing in a similar time range. The IQL baseline trains for 20\,K gradient steps with batch size 256. Across the full benchmark matrix, the main experiments (three environments, four method blocks, three seeds) were executed sequentially on this single-GPU workstation. The AntMaze extension, medium and medium-expert variant experiments, and supplementary analyses (architecture comparison, component-level update sweep, multi-attack auditing) were likewise conducted on the same machine.


\subsection{AntMaze Exploratory Extension}
\label{sec:appendix_antmaze_extension}

\input{figures/TABLE_A18_antmaze_extension}

Table\ref{tab:antmaze_extension} reports three-seed means of the AntMaze extension results together with hierarchical bootstrap intervals for the forget-set AUC. It should be interpreted as a supplementary summary rather than as a direct continuation of the locomotion benchmark. The ordering of the baselines is environment dependent. In U-Maze, the five baselines remain relatively close on the privacy-side auxiliary quantity, whereas U-Maze-Diverse and Medium-Diverse show larger separation in forget-set AUC under a much lower utility scale. These results therefore extend the empirical scope of the audit to a navigation family with sparse-reward difficulty, but they do not establish a score-equivalent ranking relative to the locomotion benchmark.
\section{Architecture Comparison and Controls}
\label{sec:appendix_arch_comparison}

\subsection{Cross-Architecture Configuration}
\label{sec:appendix_arch}

To compare model architectures under a shared training configuration, the same nominal optimization budget is maintained across all three paradigms. \Cref{tab:architecture} presents the complete TMI comparison under this shared training configuration.

\input{figures/TABLE_A1_architecture}

\begin{figure}[t]
	\centering
	\includegraphics[width=0.6\linewidth]{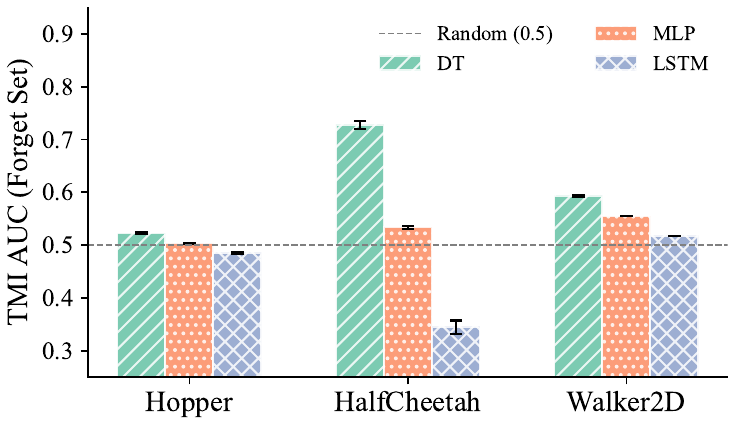}
	\caption{Architecture comparison for the replay variants under the shared training configuration. Bars report the mean forget-set AUC across seeds for DT, MLP, and LSTM. This figure complements \Cref{tab:architecture} by giving a compact visual summary of the architecture-associated ranking, which remains descriptive rather than capacity-matched.}
	\label{fig:appendix_arch_comparison}
\end{figure}

All architectures (DT, LSTM, and MLP) are trained for exactly $100$K gradient steps using a context length of $K=20$ timesteps, where applicable. Parameter counts vary materially, with the Decision Transformer ($\sim$727K) having higher capacity than the LSTM ($\sim$594K) and the MLP ($\sim$69$--$71K). Consequently, the ordering of membership scores should be interpreted as a shared-budget comparison rather than a parameter-matched statement regarding architecture alone. As demonstrated in \Cref{tab:architecture}, the MLP achieves mean D4RL normalized scores of $15.28$ in Hopper, $20.46$ in Walker2D, and $35.32$ in HalfCheetah across seeds $0$, $1$, and $2$. Here, locomotion utility follows the standard D4RL normalization, where higher values indicate better task performance on a scale aligned with the benchmark reference policies. The LSTM contains approximately eight times more parameters than the MLP, yet its forget gap is larger in Hopper and HalfCheetah and smaller in Walker2D, so the shared-budget comparison does not support a uniform cross-environment ordering between these two baselines.

The main Decision Transformer configuration uses three layers, a hidden dimension of $128$, a single attention head per layer, and dropout of $0.1$, together with a Gaussian action head. Optimization uses AdamW with batch size $64$, learning rate $10^{-4}$, weight decay $10^{-4}$, linear warmup for $10$K steps, and gradient clipping at $0.25$. The LSTM baseline also uses three layers with embedding dimension $128$ and dropout $0.1$, while the MLP baseline uses two hidden layers of width $256$ with dropout $0.1$. The IQL configuration uses two hidden layers of width $256$, dropout $0.0$, batch size $256$, $20$K gradient steps, discount $0.99$, expectile $0.7$, temperature $3.0$, and Adam-based actor, critic, and value learning rates of $3\times10^{-4}$.

Across the main benchmark tables, results are aggregated over random seeds $0$, $1$, and $2$ unless stated otherwise. The experiments are implemented in Python $3.12$, PyTorch $2.2$, and Gymnasium with MuJoCo support version $1.2.3$.

\subsection{Within-Family Capacity Control and Cross-Family Anchors}
\label{sec:appendix_architecture_control}

To mitigate the influence of potential confounds in the architecture comparison under a shared training configuration, two additional analyses are provided. The first analysis remains within the DT family to compare effective-capacity variants under an identical training horizon. The second analysis adds cross-family anchors in Hopper, HalfCheetah, and Walker2D by pairing medium DT checkpoints with MLP comparison runs from the same scaling settings, and by introducing closer parameter-matched MLP controls in Hopper and Walker2D. These perspectives reduce the capacity mismatch and test whether the reported differences vanish under a stronger control.

\input{figures/TABLE_A3_architecture_control}

\Cref{tab:architecture_control} demonstrates that the within-family DT comparisons retain nontrivial forget-gap variation even before extending across model families. In Hopper, the medium DT reaches a larger forget gap than the small DT at higher utility, whereas the medium-to-large transition fails to yield a monotone capacity-only pattern once utility deteriorates sharply. HalfCheetah provides only a noisier same-family check: the medium DT retains a larger mean forget gap than the large DT, but this setting remains low power, and the utility difference is still substantial. Walker2D exhibits an equally weak monotone trend: the large DT improves utility relative to the medium DT, yet the forget gap changes only from $0.125$ to $0.120$. Therefore, these within-family controls support only a residual architecture-associated signal rather than a clean capacity law.

The cross-family anchors in the lower panels of \Cref{tab:architecture_control} indicate a similar direction while remaining explicitly limited in scope. The approximate anchors preserve the earlier pattern: in Hopper, HalfCheetah, and Walker2D, the medium DT retains a larger forget gap than the paired MLP anchor, but the utility mismatch remains substantial, especially in HalfCheetah and Walker2D. The closer parameter-matched controls strengthen this reading in a narrower sense. In Hopper, the parameter-matched MLP still exhibits a smaller mean forget gap than the paired medium DT, while also retaining lower mean utility. In Walker2D, the parameter-matched control brings the mean forget gaps much closer, yet the paired MLP remains weaker in utility. The table therefore reduces the original ten-fold capacity concern without overturning the conservative conclusion of the supplement: the shared-budget comparison is not explained solely by the initial parameter gap, yet the remaining differences in sequence modeling, return conditioning, and optimization still prevent a clean causal claim about architecture alone.

\input{figures/TABLE_A16_context_length_control}

The same-family context-length check in \Cref{tab:context_length_control} provides a limited mechanism probe for the medium DT configuration. In Hopper, the forget-set AUC remains nearly unchanged across context lengths $K \in \{10, 20, 40\}$, while the D4RL normalized score varies substantially. Walker2D shows the clearest utility sensitivity, with the D4RL normalized score rising at $K{=}20$ before partially declining at $K{=}40$, whereas the forget-set AUC stays near $0.52$ throughout and the mean interval width remains stable at approximately $0.185$ to $0.186$. HalfCheetah remains dominated by low statistical power: the mean AUC stays close to random guessing, and the mean interval width remains wide at $0.435$ to $0.439$ across all three context lengths. These additional runs therefore do not support a simple monotone account in which longer temporal context alone produces stronger membership leakage. Instead, they reinforce the narrower interpretation adopted throughout the paper, namely that utility, architecture, and audit behavior interact in an environment-dependent manner even within a fixed DT family.

\input{figures/TABLE_A17_architecture_matched_control}

\Cref{tab:architecture_matched_control} consolidates the same-family DT controls, the approximate DT-versus-MLP anchors, and the IQL companion readout into a single matched-control summary. The common pattern is that the medium DT retains a larger forget gap than the closest available anchors in Hopper, while HalfCheetah remains the clearest reminder of the remaining limitation because every comparison in that environment is filtered through a low-power regime with only approximate controls. Walker2D further shows that the architecture signal is sensitive to the choice of control, since the gap relative to the approximate MLP anchor is modest even before introducing the tighter parameter-matched comparison discussed above. Consequently, the strengthened supplement narrows the interpretation of the architecture evidence rather than broadening it: the current experiments reduce the force of the original capacity objection, but they still support only an architecture-associated signal under the evaluated configurations, not architecture as the unique driver of membership leakage.

\subsection{Non-DT Baseline Supplement}
\label{sec:appendix_non_dt_baselines}

To evaluate non-DT architectures, a shared-training-configuration comparison is presented alongside an exploratory cross-family path and the detailed IQL companion results referenced in the main text.

\input{figures/TABLE_3_iql_main_track}

\input{figures/TABLE_A7_non_dt_supplement}

The shared-training-configuration comparison in \Cref{tab:architecture} establishes that the DT attains the largest forget gap in Hopper and HalfCheetah, whereas the ordering between the MLP and the LSTM depends on the environment and remains statistically weak. Considered together with \Cref{tab:architecture_control}, the current supplementary evidence suggests that the DT-family signal is not reducible to a single broad utility mismatch, although the comparison still remains descriptive rather than causal. \Cref{tab:non_dt_supplement} extends this observation in two directions. The exploratory sequence-model rows show that non-DT sequence baselines are not entirely absent: LSTM checkpoints can achieve competitive utility in isolated settings, most notably Walker2D, yet the current sequence-model coverage remains too sparse and uneven to support a benchmark family with the same maturity as the DT pipeline. \Cref{tab:iql_main_track} provides the detailed IQL companion summary referenced in the main text, while the IQL rows in \Cref{tab:non_dt_supplement} place that companion next to the exploratory sequence-model evidence. Collectively, these results extend the empirical coverage beyond the DT family, although they still do not justify treating the IQL results as fully score-equivalent to the DT likelihood-based pipeline.

\section{Component-Level Diagnostic Probes}
\label{sec:appendix_selective_unlearning}

\subsection{Random-Mask Controls for Component-Level Updates}
\label{sec:appendix_random}

The component-level results (\Cref{sec:selective}) indicate that update sensitivity varies across DT components. To investigate whether these differences simply result from updating fewer parameters, random-mask controls are executed. For each targeted attention layer, three independent random parameter masks of identical size (approximately 66K parameters, or $9.1\%$ of the model body) are generated, and the same gradient ascent protocol is applied.

\input{figures/TABLE_A8_random_control}

The results in \Cref{tab:random_control} support the existence of environment-dependent component effects. In HalfCheetah, the targeted attention layer achieves a significantly smaller forget gap than uniform GA, whereas random masks of equivalent size remain ineffective ($p=2.6\times 10^{-5}$). This outcome indicates a component-associated effect under the replay sweep, not a general rule for target selection. Hopper shows a different pattern: the targeted layer still improves more consistently than the size-matched random masks, but the gap between the two is much smaller, so the evidence there is better interpreted as a relative-budget success case than as equally strong mechanistic evidence.

\subsection{Full Component-Level GA and Feedforward-versus-Attention Comparison}
\label{sec:appendix_selective}

\Cref{tab:full_selective} provides supplementary results for the replay component-level sweep over the uniform baseline, all attention layers jointly, and individual attention layers. In Hopper and HalfCheetah, some individual attention layers attain smaller forget gaps than the corresponding all-attention update. Walker2D is less consistent: the all-attention update is weaker than every individual layer at $100$ and $500$ steps, but one all-attention setting at $250$ steps slightly outperforms the individual-layer rows. This finding supports the main-text interpretation that component-level gains are environment dependent rather than uniform across settings.

\input{figures/TABLE_A2_full_selective}

\begin{figure}[t]
	\centering
	\includegraphics[width=0.5\linewidth]{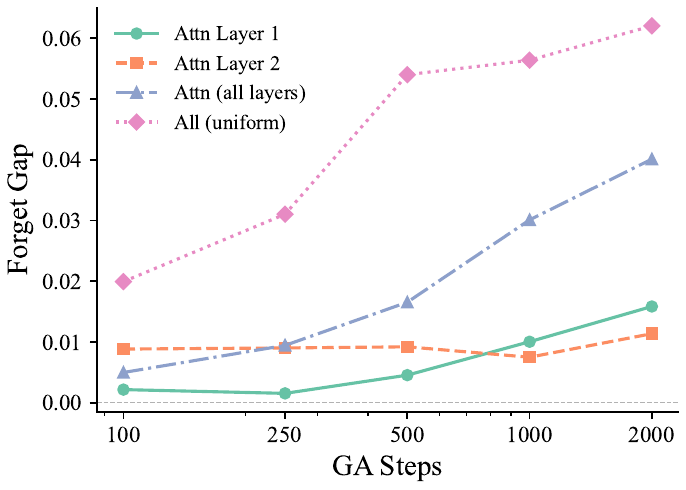}
	\caption{Hopper replay comparison between component-level and uniform GA across ascent-step budgets. The plot highlights the strongest replay-setting slice discussed in the main text and complements \Cref{tab:full_selective} with a step-by-step view of how the Hopper forget gap changes across targets. It should be interpreted as a single-environment diagnostic slice rather than as evidence for a general selective-update method.}
	\label{fig:appendix_selective_hopper}
\end{figure}

\Cref{tab:hopper_mlp_selective} turns the supplementary Hopper feedforward sweep into a matched single-environment comparison against the attention-targeted sweep under the same $\lambda=1.0$ setup and the same ascent-step budgets. Here, FFN Layer 0, FFN Layer 1, and FFN Layer 2 denote the feedforward sublayers inside the DT blocks rather than the standalone MLP baseline from \Cref{sec:models}. Feedforward-targeted updates can outperform the same-step uniform baseline once the ascent budget reaches $250$ or $500$ steps. However, the best Hopper feedforward setting, FFN Layer 0 at $250$ steps with forget gap $0.025$, remains materially weaker than the matched attention target at the same step count, namely Attention Layer 1 with forget gap $0.002$. This comparison therefore addresses the most direct target-family question supported by the current evidence: feedforward-targeted component updates are feasible in Hopper, but the strongest matched result in that environment still arises from attention targeting.

\input{figures/TABLE_A13_hopper_mlp_selective}

\subsection{Formal Utility Criteria for Component-Level Evaluation}
\label{sec:appendix_selective_criteria}

For the component-level analysis, let $U(\cdot)$ denote the evaluation score, which is the D4RL normalized score for locomotion and the task success rate for AntMaze. For a component target $c$ and the matched uniform baseline $u$ evaluated at the same ascent budget, let $\pi_c$ and $\pi_u$ denote the corresponding updated policies. The relative utility budget is defined as
\begin{equation}
	\Delta U_{\mathrm{rel}}(c \mid u) = U(\pi_c) - U(\pi_u) \geq 0,
	\label{eq:relative_utility_budget}
\end{equation}
so a component target is feasible when it does not underperform the matched uniform update. To distinguish this relative criterion from strong absolute retention, the benchmark also reports retained performance ratios for an updated policy $\pithetap$, measured against the retraining reference $\piretrain$ and the original base policy $\pitheta$:
\begin{equation}
	\rho_{\mathrm{retrain}}(\pithetap) = \frac{U(\pithetap)}{U(\piretrain)}, \qquad
	\rho_{\mathrm{base}}(\pithetap) = \frac{U(\pithetap)}{U(\pitheta)}.
	\label{eq:retained_performance_ratios}
\end{equation}
These ratios define the explicit retained-performance criteria used throughout the component-level analysis.

\subsection{Utility-Constrained Component-Level Comparison}
\label{sec:appendix_selective_budget}

To evaluate component-level updates under explicit utility constraints, the held-out targets from \Cref{tab:layer_cv} are compared against the uniform \textbf{All} baseline at matched ascent steps. A comparison is considered feasible if the D4RL normalized-score drop of the selected target relative to the uniform baseline remains within the specified budget. This relative utility criterion is intentionally weaker than the explicit retained-performance criteria reported in \Cref{tab:oracle}, so both views are required for interpretation.

\input{figures/TABLE_A12_selective_budget}

As discussed in \Cref{sec:selective} and quantified in \Cref{tab:selective_budget}, the replay component-level sweep provides relative-budget evidence for environment-dependent component sensitivity. In this replay slice, Hopper shows the largest improvement over the matched uniform baseline under the weaker criterion, while HalfCheetah improves the forget gap only in a low-power regime and Walker2D fails to demonstrate improvement for budgets up to $2.0$. Considered together with \Cref{tab:oracle} and the retained-utility check, these results indicate that relative success against the matched uniform baseline does not by itself imply satisfaction of the explicit retained-performance criteria. The overall evidence therefore supports the main-text interpretation that component-level updating is most useful as a diagnostic probe of environment-dependent structure rather than as a general unlearning method.

\begin{figure}[t]
	\centering
	\includegraphics[width=\textwidth]{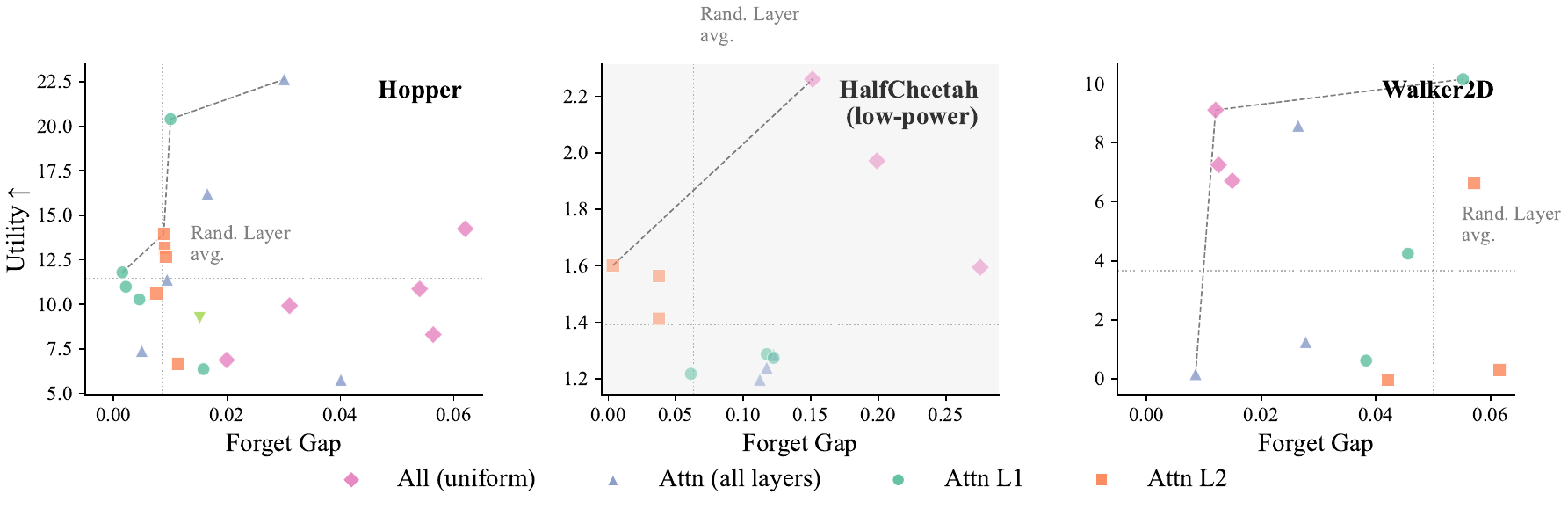}
	\caption{Privacy-utility tradeoff across GA targets and step counts. Component-level updates can improve relative forget gaps in some settings, while the broader frontier shows that smaller gaps do not necessarily imply strong retained utility.}
	\label{fig:pareto}
\end{figure}

\subsection{Pre-Unlearning Gradient Diagnostics}
\label{sec:appendix_localization}

To complement the post-hoc component-level sweep, the benchmark logs a pre-unlearning component-sensitivity diagnostic before the first update of the $\lambda=1.0$, $500$-step GA+Refit runs. For each candidate component, the ratio between gradient norms on pre-update minibatches from the forget set and the retain set is computed using two shared minibatches.

\input{figures/TABLE_A14_localization_formal}

\Cref{tab:localization_formal} shows that these ratios are consistently non-uniform across environments, but the dominant component is not always an attention layer. HalfCheetah is uniformly dominated by FFN Layer 2, Hopper most often by FFN Layer 0, and Walker2D by FFN Layer 1, matching the component names reported in the table. The mean top ratios range from $1.182$ to $2.071$, indicating that the strongest forget gradients can substantially exceed the corresponding retain gradients. These diagnostics therefore support heterogeneous component sensitivity, while also explaining why they are not used as a direct target-selection rule. The dominant component depends on the environment, and a large pre-update gradient ratio does not by itself imply that the same component yields the best post-update privacy-utility evidence. In HalfCheetah, this distinction is visible directly in the data, because the gradient diagnostic concentrates on a feedforward component while the strongest component-level update under the evaluated budget occurs in Attention Layer 2.

\section{Matching and Statistical Validation}
\label{sec:appendix_matching_stat}

\subsection{Matching Diagnostics and Precision Audit}
\label{sec:appendix_matching_power}

\input{figures/TABLE_A4_matching_power}

\Cref{tab:matching_power} makes the role of matching explicit in the selected high-utility locomotion settings used for the main multi-attack audit. One-to-one pairing succeeds for every forget trajectory, and the largest confounds in return and trajectory length decrease after matching. In Walker2D medium, the absolute standardized mean difference for return decreases from $0.346$ to $0.003$, and the corresponding length imbalance decreases from $0.334$ to $0.005$. Walker2D medium-expert and HalfCheetah medium-expert exhibit the same qualitative pattern, with return imbalance decreasing from $0.041$ to $0.004$ and from $0.078$ to $0.002$, respectively. The same table also quantifies the residual initial-state mismatch. The post-match $s_0$-norm absolute SMD is $0.082$ in Walker2D medium, $0.045$ in Walker2D medium-expert, and $0.020$ in HalfCheetah medium-expert, while the post-match maximum coordinate-level $s_0$ absolute SMD remains $0.365$, $0.237$, and $0.207$, respectively. The matching procedure should therefore be interpreted as reducing coarse observable confounds rather than as proving full covariate balance.

The same table summarizes the effective precision of these selected high-utility settings under a stricter uncertainty analysis. Walker2D medium-expert is the most stable case in this table, with a hierarchical interval of $[0.499, 0.534]$, mean paired-bootstrap interval width $0.060$, and approximate detectable gap $0.093$. HalfCheetah medium-expert remains less precise, with interval $[0.609, 0.672]$, mean paired width $0.110$, and detectable gap $0.097$, while Walker2D medium sits between them at interval $[0.591, 0.651]$, mean paired width $0.104$, and detectable gap $0.126$. Therefore, even within this higher-utility comparative slice, precision remains environment dependent. The broader power analysis reported elsewhere in our benchmark artifacts further indicates that some replay settings, especially HalfCheetah medium-replay, operate in substantially lower-power regimes.

\input{figures/TABLE_A15_matching_variants}

\Cref{tab:matching_variant} extends this robustness check from the base DT to the full canonical medium-replay method set in Hopper and Walker2D. The stronger feature-matched variant augments the original matching rule with reward, action, and trajectory-dynamics summaries while preserving one-to-one pairing. In Walker2D, this stricter matching consistently reduces the mean forget gap across all four benchmark blocks, from $0.125$ to $0.065$ for the base DT, from $0.020$ to $0.008$ for the retraining reference, from $0.028$ to $0.005$ for naive fine-tuning, and from $0.089$ to $0.054$ for GA+Refit. Hopper shows a weaker and less uniform sensitivity: the stronger variant slightly increases the gaps for the base DT, retraining reference, and naive fine-tuning, but each of those values remains close to random guessing, whereas the GA+Refit gap decreases from $0.052$ to $0.030$. The unmatched baseline provides a more severe stress test. In Walker2D, removing matching increases the mean forget gap for three of the four benchmark blocks and leaves the base DT nearly unchanged, while also causing the GA+Refit runs to fail the validity check for all three seeds. In Hopper, the unmatched baseline pushes the base model, retraining reference, and naive fine-tuning even closer to random guessing, yet it also destabilizes the GA+Refit slice, where one seed fails the validity check and the remaining two sit only marginally above the rejection threshold. These comparisons support a narrower interpretation of the replay-setting evidence: the measured forget gaps do depend on the construction of the matched non-member set, but the canonical matched benchmark remains a conservative control rather than an artifact that creates the method ordering by itself.

\subsection{Quartile Analysis and Reward-Dependent Memorization}
\label{sec:appendix_quartile}

To identify which data cohorts are most susceptible to privacy leaks, reward-dependent memorization is analyzed by partitioning the forget set into quartiles (Q1--Q4) based on episode return.

\begin{figure}[t]
	\centering
	\includegraphics[width=0.6\textwidth]{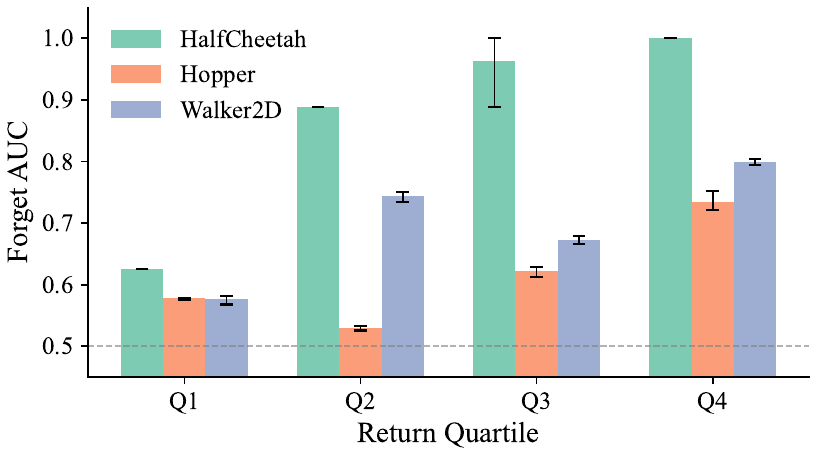}
	\caption{TMI AUC across trajectory return quartiles for HalfCheetah, Hopper, and Walker2D. High-return trajectories often exhibit the strongest initial baseline membership signals, although the quartile ordering is not identical across environments.}
	\label{fig:quartile}
\end{figure}

In HalfCheetah and Hopper, the upper-return trajectories provide the clearest initial privacy risk, while Walker2D follows a different pattern. In HalfCheetah, both Q3 and Q4 attain mean baseline AUC values close to $1.0$, with Q4 remaining highest. The full quartile decay analysis further shows that this Q4 advantage attenuates under extended naive fine-tuning, with the mean HalfCheetah Q4 AUC decreasing from $1.0$ at step $0$ to $0.625$ at $10^5$ steps, while remaining above Q3 throughout. Walker2D instead shows a disproportionately high baseline membership signal for Q2 (AUC $0.744$). This contrast indicates that reward-dependent memorization is not governed solely by the total episode return.

\begin{figure}[t]
	\centering
	\includegraphics[width=0.98\textwidth]{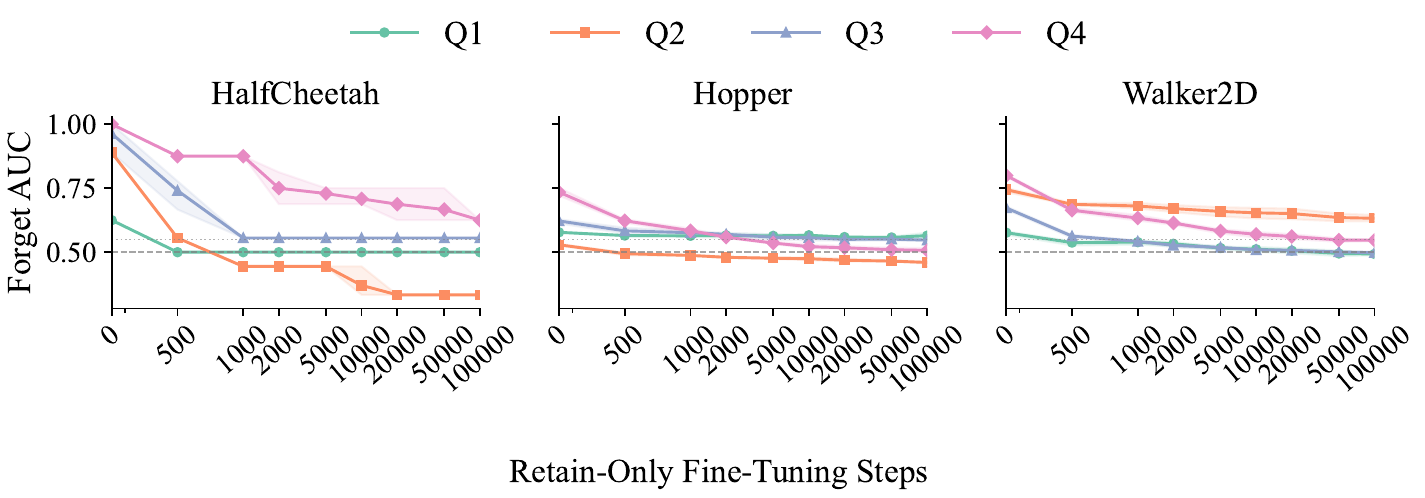}
	\caption{Cross-environment quartile decay trajectories under naive fine-tuning. Each panel shows seed means with seed-range bands. Persistence is strongly environment dependent: HalfCheetah retains the two upper quartiles longest, Walker2D preserves elevated signals for Q2 and Q4, and Hopper shows a weaker separation in which the initial Q4 advantage does not remain dominant at long budgets.}
	\label{fig:quartile_decay_envs}
\end{figure}

The HalfCheetah panel of \Cref{fig:quartile_decay_envs} makes the persistence structure explicit. The lower-return quartiles are erased quickly, but the two upper quartiles remain harder to suppress throughout the full naive fine-tuning budget. This separation clarifies why the baseline bar chart in \Cref{fig:quartile} should be interpreted as an initial ordering rather than a claim of static privacy risk. HalfCheetah preserves the ranking $\text{Q4} > \text{Q3} > \text{Q2} > \text{Q1}$ only approximately at the start, after which Q4 still remains the most persistent cohort while its margin over Q3 narrows.

The cross-environment comparison in \Cref{fig:quartile_decay_envs} shows that reward-dependent memorization does not translate into a uniform persistence rule. HalfCheetah provides the clearest monotone ordering, with Q1 and Q2 crossing the $0.55$ threshold by $500$ and $1000$ steps, while Q3 and Q4 remain above that threshold throughout the full budget. Walker2D exhibits a different pattern in which Q2 remains persistently elevated and Q4 decays more slowly than Q1 and Q3. Hopper is less separable still: Q4 starts as the strongest cohort at baseline, yet by long budgets its mean AUC falls near random-guessing levels while Q1 remains slightly above the other quartiles. The benchmark therefore supports a qualified conclusion: higher-return trajectories are the clearest initial privacy risk, but the long-horizon persistence of that risk is environment specific.

\subsection{Statistical Rigor: TOST Sensitivity Analysis}
\label{sec:appendix_tost}

To assess the robustness of the privacy claims, a sensitivity analysis is performed using Two One-Sided Tests (TOST) under increasingly stringent equivalence margins $\varepsilon$. The main text reports the benchmark-default margin $\varepsilon = 0.1$, whereas this appendix adds a stricter follow-up check at $\varepsilon = 0.05$. The test asks whether the forget-set AUC is statistically equivalent to random guessing, namely to $0.5$, within the specified margin.

\input{figures/TABLE_A9_tost_sensitivity}

\Cref{tab:tost_sensitivity} demonstrates that at $\varepsilon = 0.05$, the highlighted component-level configurations in the replay variants of Hopper, HalfCheetah, and Walker2D all pass the equivalence test. Under this margin, passing corresponds to statistical equivalence to random guessing within an AUC band of $0.5 \pm 0.05$, which matches the $0.55$ threshold used in the quartile-decay discussion above. The highlighted rows are tied to the cross-validated target family in \Cref{tab:layer_cv}, while the ascent step is selected among candidates that also satisfy the matched-baseline condition in the stability analysis. These replay-setting results show that component-level GA can approach random-guessing-level privacy under strict equivalence criteria, while retained-performance evidence still depends on the stronger utility checks reported elsewhere.

\subsection{Cross-Validation of Layer Selection}
\label{sec:appendix_cv}

To verify that the identified layers are not artifacts of specific random seeds, leave-one-seed-out cross-validation is performed in the replay component-level sweep. The results in \Cref{tab:layer_cv} show that stable target choices vary by environment and that cross-validated transfer under a relative utility budget is weaker than satisfying explicit retained-performance criteria. This pattern reinforces the conclusion that component targets are environment-dependent diagnostic signals rather than a general update rule.

\input{figures/TABLE_A10_layer_cv}

\section{Multi-Attack Privacy Auditing}
\label{sec:appendix_multi_attack}

\subsection{Action-Error Membership Attack}
\label{sec:appendix_action_error}

To complement the likelihood-derived audit, a membership attack based on action error is evaluated. For each trajectory, the mean squared error (MSE) between the predicted action mean and the observed action is computed. This non-likelihood score family serves as an additional robustness check.

\input{figures/TABLE_A5_action_error}

\Cref{tab:action_error_attack} demonstrates that this score family fails to yield a uniform ordering across the high-utility locomotion settings. In HalfCheetah medium-expert, the base DT retains a clear action-error membership signal, while the retraining reference and naive fine-tuning reduce it more strongly than GA+Refit or TrajDeleter. Walker2D medium-expert stays close to random guessing for all methods, whereas Walker2D medium separates the base DT, GA+Refit, and TrajDeleter from the retraining and fine-tuning references. Consequently, the overall pattern is consistent with the main text: stronger deletion pressure does not reliably deliver a superior privacy-utility profile compared to simpler baselines.

\subsection{Representation-Based Membership Attack}
\label{sec:appendix_representation_attack}

A representation-based attack is further included, where a logistic-regression classifier is trained in cross-validation to predict membership from averaged final-layer state-token representations. This approach provides an alternative perspective independent of likelihood or prediction error.

\input{figures/TABLE_A6_representation_shadow}

The upper panel of \Cref{tab:representation_shadow_attack} illustrates that the relative ranking of methods is attack-dependent. The representation-based attack is weak and unstable in most high-utility settings, with many AUC values close to random guessing or below $0.5$. Values below $0.5$ indicate an inverted ranking in which the learned representation makes matched non-members appear easier than forget trajectories under this classifier, rather than providing evidence of stronger deletion. This instability does not provide an independent method ordering, but it motivates the necessity of the multi-attack protocol used throughout the study.

\subsection{Query-Limited Shadow-Model Attack}
\label{sec:appendix_shadow_query_attack}

To provide a practical black-box perspective, a query-limited shadow-model attack is evaluated. Eight time steps are sampled per trajectory, the model is queried for action means, and a membership classifier is trained on the resulting action errors using shadow outputs.

The lower panel of \Cref{tab:representation_shadow_attack} reaches the same broader conclusion from this more restricted setting. The query-limited shadow attack remains sensitive to calibration against retraining references: in Walker2D medium, for example, all methods have high absolute AUC values, including the retraining reference and naive fine-tuning. The shadow-query results therefore should not be treated as an independent verdict. They are most informative when read together with the likelihood, reference-model, action-error, and retained-utility diagnostics.

\section{Extensions}
\label{sec:appendix_extensions}

\subsection{Fisher Baseline Supplement}
\label{sec:appendix_fisher_baseline}

To connect the benchmark with Fisher-information-based unlearning methods, a diagonal Fisher baseline is evaluated under the high-utility locomotion settings used in the main multi-attack audit. The standard hyperparameter configuration uses a damping factor of $10^{-3}$, a scale of $1.0$, and head refitting. The resulting checkpoints are aggregated across seeds $0$, $1$, and $2$.

\input{figures/TABLE_A11_fisher_baseline}

\Cref{tab:fisher_baseline} clarifies the scope of the Fisher result referenced in the main text. Across the high-utility locomotion settings, Fisher updates can place the forget-set AUC near random guessing, but only with severe D4RL score collapse. The environment means remain in a low-utility regime for HalfCheetah medium-expert, Walker2D medium-expert, and Walker2D medium, far below the benchmarked retraining and fine-tuning baselines.

\begin{figure}
	\centering
	\includegraphics[width=0.5\linewidth]{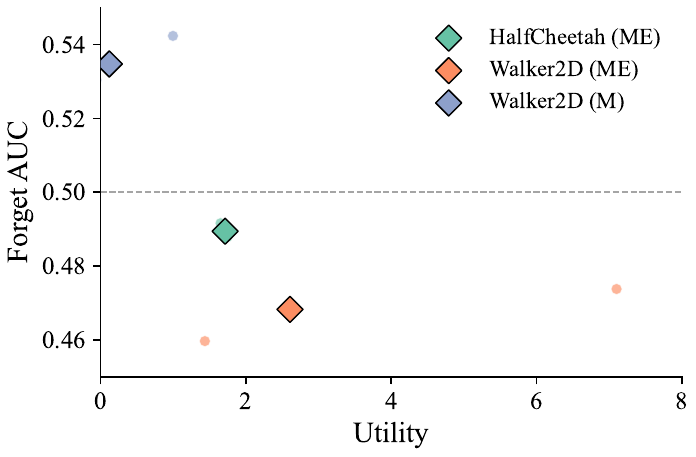}
	\caption{Privacy-utility scatter for the diagonal Fisher baseline across seeds $0$, $1$, and $2$. Each small point is one seed, and each diamond is the environment mean. Across the high-utility locomotion settings, privacy scores can approach random guessing only in a low-utility regime.}
	\label{fig:fisher_tradeoff}
\end{figure}

\Cref{fig:fisher_tradeoff} complements the table by making the cross-environment pattern more explicit. The seed-level points show that near-random privacy scores are coupled with extremely low D4RL scores, and settings with slightly higher utility do not alter this pattern. Consequently, this baseline is interpreted as supplementary evidence that curvature-aware updates do not eliminate the privacy-utility tension within this benchmark, rather than serving as a competitive alternative to the primary methods.

\subsection{Limitations}
\label{sec:appendix_limitations}

This study presents several limitations. Because the employed DT uses a single attention head per layer, the benchmark does not resolve whether membership signals differ across individual heads. The cross-architecture results also remain confounded by differences in model capacity, optimization behavior, achieved utility, and audit family, despite the shared nominal training budget and the closer anchors reported in \Cref{tab:architecture_control}. The environment coverage extends beyond MuJoCo locomotion through three AntMaze settings, but those results rely on a success-rate-derived utility scale and therefore should be interpreted as a separate navigation case study rather than as a direct continuation of the locomotion score scale. The component-level evidence is similarly uneven across environments: Walker2D medium-expert contains a narrow retained-utility signal for attention-only updates, but the same settings still fail the retain-side shift checks, and the pattern does not replicate across the other main high-utility settings. Finally, the attack suite still covers only a narrow subset of practical attackers, and statistical power remains limited in settings where matched-pair counts are small and detectable gaps are correspondingly large.

%% file: figures/TABLE_A18_antmaze_extension.tex
\begin{table}[htp]
\centering
\renewcommand{\arraystretch}{0.8}
\caption{AntMaze extension over U/UD/MD variants. Utility denotes a success-rate-derived scale rather than the D4RL normalized locomotion score used in the main benchmark table, so these values should not be compared numerically with \Cref{tab:benchmark}. Each row reports the mean over three seeds. Forget-Set AUC remains the same direction-sensitive auxiliary quantity used elsewhere in TOUR. The 95\% CI is a hierarchical bootstrap interval over the three seeds with paired resampling inside each seed. These rows extend the audit to a navigation family with a distinct utility scale.}
\label{tab:antmaze_extension}
\resizebox{0.78\textwidth}{!}{%
\begin{tabular}{l|lrrrcr}
\toprule
Environment & Method & Utility $\uparrow$ & Forget Gap $\downarrow$ & Forget AUC & 95\% CI & Retain AUC \\
\midrule
\multirow{5}{*}{AntMaze (U)} & Base DT & 53.00 & 0.012 & 0.512 & [0.433, 0.536] & 0.493 \\
 & Retrain Ref. & 63.00 & \textbf{0.004} & 0.504 & [0.397, 0.503] & 0.561 \\
 & Naive FT & 60.67 & 0.011 & 0.511 & [0.411, 0.517] & 0.566 \\
 & GA+Refit & 64.00 & 0.094 & 0.406 & [0.317, 0.422] & 0.508 \\
 & TrajDeleter & 57.33 & 0.016 & 0.516 & [0.433, 0.550] & 0.499 \\
\midrule
\multirow{5}{*}{AntMaze (UD)} & Base DT & 25.67 & 0.500 & 1.000 & [1.000, 1.000] & 1.000 \\
 & Retrain Ref. & 13.67 & \textbf{0.000} & 0.500 & [0.167, 0.833] & 1.000 \\
 & Naive FT & 7.33 & 0.167 & 0.667 & [0.167, 1.000] & 1.000 \\
 & GA+Refit & 0.00 & 0.500 & 0.000 & [0.000, 0.000] & 1.000 \\
 & TrajDeleter & 31.00 & 0.500 & 1.000 & [1.000, 1.000] & 1.000 \\
\midrule
\multirow{5}{*}{AntMaze (MD)} & Base DT & 9.33 & 0.146 & 0.646 & [0.250, 0.750] & 0.587 \\
 & Retrain Ref. & 17.67 & \textbf{0.021} & 0.479 & [0.250, 0.750] & 0.854 \\
 & Naive FT & 21.00 & 0.042 & 0.542 & [0.250, 0.750] & 0.879 \\
 & GA+Refit & 0.00 & 0.438 & 0.062 & [0.000, 0.000] & 0.896 \\
 & TrajDeleter & 9.33 & 0.167 & 0.667 & [0.250, 0.750] & 0.614 \\
\bottomrule
\end{tabular}%
}
\end{table}

%% file: figures/TABLE_A1_architecture.tex
\begin{table}[h]
\centering
\renewcommand{\arraystretch}{0.8}
\caption{Cross-architecture TMI comparison under shared training budget (100K steps, context=20). Parameter counts differ substantially (DT $\sim$727K, LSTM $\sim$594K, MLP $\sim$71K). Claims are scoped to this experimental setting and should not be read as parameter-matched architecture isolation.}
\label{tab:architecture}
\resizebox{0.6\linewidth}{!}{%
\begin{tabular}{l|lrrrc}
\toprule
Environment                  & Model & Params & Utility $\uparrow$  & Forget AUC & 95\% CI        \\
\midrule
\multirow{3}{*}{Hopper} & DT & 726K & 22.35 & 0.523 & [0.521, 0.524] \\
 & MLP & 69K & 15.28 & 0.503 & [0.503, 0.504] \\
 & LSTM & 593K & 1.70 & 0.485 & [0.483, 0.487] \\
\midrule
\multirow{3}{*}{HalfCheetah} & DT & 727K & 36.77 & 0.728 & [0.719, 0.735] \\
 & MLP & 71K & 35.32 & 0.534 & [0.531, 0.536] \\
 & LSTM & 594K & 1.81 & 0.345 & [0.332, 0.357] \\
\midrule
\multirow{3}{*}{Walker2D} & DT & 727K & 45.22 & 0.593 & [0.591, 0.595] \\
 & MLP & 71K & 20.46 & 0.555 & [0.555, 0.555] \\
 & LSTM & 594K & 0.31 & 0.518 & [0.517, 0.518] \\
\bottomrule
\end{tabular}
}
\end{table}

%% file: figures/TABLE_A3_architecture_control.tex
\begin{table}[htp]
\centering
\caption{Within-family DT capacity controls together with approximate and closer parameter-matched cross-family anchors for the medium-replay settings. The within-family rows isolate residual variation inside the DT family under the same training horizon. The cross-family rows reduce the original capacity mismatch relative to \Cref{tab:architecture}, but they still do not provide strict architecture isolation because sequence modeling, return conditioning, and optimization remain coupled.}
\label{tab:architecture_control}
\resizebox{\textwidth}{!}{%
\begin{tabular}{llrrrrr}
\toprule
Panel                                        & Environment & Model pair       & Utility $\uparrow$ & Forget AUC & Forget Gap $\downarrow$ & 95\% CI width                 \\
\midrule
\multirow{7}{*}{Within-family DT} & Hopper & S$_{128\times3}$ & 22.52 & 0.582 & \textbf{0.082} & 0.132 \\
 & Hopper & M$_{256\times3}$ & 28.52 & 0.626 & 0.127 & 0.129 \\
 & Hopper & L$_{256\times6}$ & 17.86 & 0.606 & 0.106 & 0.131 \\
 & HalfCheetah & M$_{256\times3}$ & 34.86 & 0.784 & 0.284 & 0.355 \\
 & HalfCheetah & L$_{256\times6}$ & 28.03 & 0.752 & \textbf{0.252} & 0.391 \\
 & Walker2D & M$_{256\times3}$ & 26.95 & 0.625 & 0.125 & 0.177 \\
 & Walker2D & L$_{256\times6}$ & 35.99 & 0.620 & \textbf{0.120} & 0.178 \\
\midrule
\multirow{3}{*}{Approx. cross-family anchor} & Hopper & DT(M) vs. MLP & 28.52 / 20.02 & --- & 0.127 / 0.007 & $\Delta_{\mathrm{gap}}=0.119$ \\
 & HalfCheetah & DT(M) vs. MLP & 34.86 / 19.11 & --- & 0.284 / 0.022 & $\Delta_{\mathrm{gap}}=0.262$ \\
 & Walker2D & DT(M) vs. MLP & 26.95 / 5.38 & --- & 0.125 / 0.039 & $\Delta_{\mathrm{gap}}=0.086$ \\
\midrule
\multirow{2}{*}{Parameter-matched cross-family anchor} & Hopper & DT(M) vs. MLP$_{\mathrm{pm}}$ & 28.52 / 14.59 & --- & 0.127 / 0.028 & $\Delta_{\mathrm{gap}}=0.099$ \\
 & Walker2D & DT(M) vs. MLP$_{\mathrm{pm}}$ & 26.95 / 19.61 & --- & 0.125 / 0.116 & $\Delta_{\mathrm{gap}}=0.010$ \\
\bottomrule
\end{tabular}%
}
\end{table}

%% file: figures/TABLE_A16_context_length_control.tex
\begin{table}[htp]
\centering
\caption{Context-length sensitivity of the medium DT scaling checkpoint $M_{256\times3}$ under a fixed $500$K pretraining budget. This auxiliary probe uses a larger pretraining budget than the $100$K-step main benchmark in order to study context-length effects within the same scaling family. Each row reports the multiseed mean forget-set AUC, the mean seed-level 95\% interval width for that AUC, and the Utility after re-evaluating the same model family at context lengths $10$, $20$, and $40$. This supplement is intended as a within-family mechanism check rather than as a standalone architecture comparison.}
\label{tab:context_length_control}
\resizebox{\textwidth}{!}{%
\begin{tabular}{l|p{1.5cm}p{1.2cm}p{1.2cm}p{1.2cm}p{1.2cm}p{1.2cm}p{1.2cm}p{1cm}p{1.1cm}}
\toprule
Environment & $K{=}10$ Forget AUC & $K{=}10$ CI Width & $K{=}10$ Utility & $K{=}20$ Forget AUC & $K{=}20$ CI Width & $K{=}20$ Utility & $K{=}40$ Forget AUC & $K{=}40$ CI Width & $K{=}40$ Utility \\
\midrule
Hopper & 0.480 & 0.132 & 29.33 & 0.479 & 0.133 & 15.84 & 0.481 & 0.133 & 29.73 \\
HalfCheetah & 0.491 & 0.439 & 35.44 & 0.481 & 0.437 & 35.64 & 0.481 & 0.435 & 36.25 \\
Walker2D & 0.520 & 0.185 & 28.76 & 0.520 & 0.185 & 35.98 & 0.522 & 0.186 & 32.54 \\
\bottomrule
\end{tabular}%
}
\end{table}

%% file: figures/TABLE_A17_architecture_matched_control.tex
\begin{table}[htp]
\centering
\caption{Consolidated matched-control summary for the replay-setting architecture evidence. The table combines the medium DT anchor, the closest within-family DT comparison available for each environment, the approximate DT-versus-MLP anchor, and the IQL family companion readout. The IQL entries are reported only as a trajectory-level action-error companion audit and not as a score-equivalent replacement for the likelihood-based DT audit.}
\label{tab:architecture_matched_control}
\resizebox{\textwidth}{!}{%
\begin{tabular}{l|p{1.5cm}p{1.2cm}p{1.2cm}p{1.2cm}p{1.2cm}p{1.2cm}p{1.2cm}p{1cm}p{1.1cm}}
\toprule
Environment & DT(M) Utility & DT(M) Forget Gap & DT(L) Utility & DT(L) Forget Gap & MLP Utility & MLP Forget Gap & IQL Utility & IQL Forget Gap \\
\midrule
Hopper & 28.52 & 0.127 & 17.86 & 0.106 & 20.02 & 0.007 & 44.42 & 0.018 \\
HalfCheetah & 34.86 & 0.284 & 28.03 & 0.252 & 19.11 & 0.022 & 39.10 & 0.068 \\
Walker2D & 26.95 & 0.125 & 35.99 & 0.120 & 5.38 & 0.039 & 67.79 & 0.085 \\
\bottomrule
\end{tabular}%
}
\end{table}

%% file: figures/TABLE_3_iql_main_track.tex
\begin{table}[htp]
\centering
\renewcommand{\arraystretch}{0.8}
\caption{Primary non-DT companion summary for the replay settings. The table reports the IQL family under the trajectory-level action-error membership audit, which complements the DT-centered likelihood audit in \Cref{tab:benchmark} rather than replacing it. IQL does not parameterize the Gaussian action distribution needed for exact per-token NLL evaluation, so the action-error audit is the closest trajectory-level companion available in the current benchmark. Hopper and Walker2D provide the primary comparative settings. HalfCheetah (R) is retained as a lower-precision reference with only 14 matched pairs.}
\label{tab:iql_main_track}
\resizebox{0.9\textwidth}{!}{%
\begin{tabular}{l|lrrrr}
\toprule
Environment & Method & Utility $\uparrow$ & Forget AUC & Forget Gap $\downarrow$ & Retain AUC \\
\midrule
\multirow{4}{*}{Hopper} & IQL Base & 44.42 & 0.518 & 0.018 & 0.564 \\
 & IQL Retraining Reference & 45.65 & 0.502 & \textbf{0.002} & 0.585 \\
 & IQL Naive Fine-Tuning & 52.46 & 0.507 & 0.007 & 0.595 \\
 & IQL Actor-Ascent+Refit & 39.48 & 0.510 & 0.010 & 0.581 \\
\midrule
\multirow{4}{*}{Walker2D} & IQL Base & 67.79 & 0.585 & 0.085 & 0.508 \\
 & IQL Retraining Reference & 61.60 & 0.523 & \textbf{0.023} & 0.520 \\
 & IQL Naive Fine-Tuning & 69.90 & 0.542 & 0.042 & 0.551 \\
 & IQL Actor-Ascent+Refit & 64.70 & 0.549 & 0.049 & 0.533 \\
\midrule
\multirow{4}{*}{HalfCheetah} & IQL Base & 39.10 & 0.568 & 0.068 & 0.530 \\
 & IQL Retraining Reference & 22.78 & 0.478 & 0.022 & 0.513 \\
 & IQL Naive Fine-Tuning & 38.30 & 0.522 & 0.022 & 0.554 \\
 & IQL Actor-Ascent+Refit & 31.70 & 0.510 & \textbf{0.010} & 0.509 \\
\bottomrule
\end{tabular}%
}
\vspace{-0.5cm}
\end{table}

%% file: figures/TABLE_A7_non_dt_supplement.tex
\begin{table}[htp]
\centering
\caption{Supplementary non-DT evidence beyond the shared-budget comparison in \Cref{tab:architecture}. The upper panel summarizes exploratory LSTM and Trajectory Transformer~\citep{janner2021offline} runs. The lower panel summarizes the primary IQL family under the available action-error audit. Hopper and Walker2D provide the main comparative settings. HalfCheetah is retained as a low-power reference setting.}
\label{tab:non_dt_supplement}
\resizebox{\textwidth}{!}{%
\begin{tabular}{l|llrrrr}
\toprule
Panel                                           & Environment & Model or method        & Seeds & Utility $\uparrow$ & Forget AUC   & Forget Gap $\downarrow$ \\
\midrule
\multirow{5}{*}{Exploratory sequence baselines} & HalfCheetah & LSTM & 1 & 32.08 & 0.546 & 0.046 \\
 & HalfCheetah & Trajectory Transformer & 1 & 1.75 & 0.372 & 0.128 \\
 & Hopper & LSTM & 2 & 8.79 & 0.486 & 0.014 \\
 & Hopper & Trajectory Transformer & 2 & 1.29 & 0.482 & 0.024 \\
 & Walker2D & LSTM & 1 & 63.22 & 0.539 & 0.039 \\
\midrule
\multirow{9}{*}{IQL primary summary} & HalfCheetah & IQL Base & 3 & 39.10 & 0.568 & \textbf{0.068} \\
 & Hopper & IQL Base & 3 & 44.42 & 0.518 & 0.018 \\
 & Hopper & IQL Retraining Reference & 3 & 45.65 & 0.502 & \textbf{0.002} \\
 & Hopper & IQL Naive Fine-Tuning & 3 & 52.46 & 0.507 & 0.007 \\
 & Hopper & IQL Actor-Ascent+Refit & 3 & 39.48 & 0.510 & 0.010 \\
 & Walker2D & IQL Base & 3 & 67.79 & 0.585 & 0.085 \\
 & Walker2D & IQL Retraining Reference & 3 & 61.60 & 0.523 & \textbf{0.023} \\
 & Walker2D & IQL Naive Fine-Tuning & 3 & 69.90 & 0.542 & 0.042 \\
 & Walker2D & IQL Actor-Ascent+Refit & 3 & 64.70 & 0.549 & 0.049 \\
\bottomrule
\end{tabular}%
}
\end{table}

%% file: figures/TABLE_A8_random_control.tex
\begin{table}[h]
\centering
\caption{Comparative analysis of random-mask controls versus attention-layer selective GA ($3$ seeds $\times$ $3$ masks where applicable). In HalfCheetah, targeting Attn L2 produces a much smaller forget gap than size-matched random masks, although both settings have poor utility. In Hopper, the targeted Attn L1 remains more consistent than random masks in reducing membership signals.}
\label{tab:random_control}
\resizebox{\textwidth}{!}{%
\begin{tabular}{l|llrrrr}
\toprule
Env & Target           & Params & Steps & Forget AUC     & Forget Gap $\downarrow$ & Utility $\uparrow$ \\
\midrule
\multirow{9}{*}{Hopper} & Attn~L1 & 66K & 100 & 0.500 & 0.002 & 10.99 \\
 & Random (matched) & 66K & 100 & 0.449 & 0.054 & 17.27 \\
 & Attn~L1 & 66K & 250 & 0.499 & 0.002 & 11.79 \\
 & Random (matched) & 66K & 250 & 0.423 & 0.078 & 18.90 \\
 & Attn~L1 & 66K & 500 & 0.495 & 0.005 & 10.28 \\
 & Random (matched) & 66K & 500 & 0.420 & 0.080 & 12.31 \\
 & All (uniform) & 726K & 100 & 0.480 & 0.020 & 6.88 \\
 & All (uniform) & 726K & 250 & 0.469 & 0.031 & 9.93 \\
 & All (uniform) & 726K & 500 & 0.446 & 0.054 & 10.86 \\
\midrule
\multirow{9}{*}{HalfCheetah} & Attn~L2 & 66K & 100 & 0.537 & 0.037 & 1.56 \\
 & Random (matched) & 66K & 100 & 0.311 & 0.189 & 9.20 \\
 & \textbf{Attn~L2} & 66K & 250 & \textbf{0.497} & \textbf{0.003} & 1.60 \\
 & Random (matched) & 66K & 250 & 0.299 & 0.201 & 5.88 \\
 & Attn~L2 & 66K & 500 & 0.463 & 0.037 & 1.41 \\
 & Random (matched) & 66K & 500 & 0.295 & 0.205 & 3.12 \\
 & All (uniform) & 727K & 100 & 0.349 & 0.151 & 2.26 \\
 & All (uniform) & 727K & 250 & 0.301 & 0.199 & 1.97 \\
 & All (uniform) & 727K & 500 & 0.224 & 0.276 & 1.59 \\
\bottomrule
\end{tabular}%
}
\end{table}

%% file: figures/TABLE_A2_full_selective.tex
\begin{table}
\centering
\renewcommand{\arraystretch}{0.8}
\caption{Full selective GA results across environments, targets, and step counts (3-seed mean). Forget gap = $|\AUCf-0.5|$.}
\label{tab:full_selective}
\resizebox{0.7\textwidth}{!}{%
\begin{tabular}{l|lrrrrr}
\toprule
Environment & Target & Steps & Forget AUC & Forget Gap $\downarrow$ & Utility & Retain AUC \\
\midrule
\multirow{30}{*}{Hopper} & All & 100 & 0.480 & 0.020 & 6.9 & 0.540 \\
 & All & 250 & 0.469 & 0.031 & 9.9 & 0.531 \\
 & All & 500 & 0.446 & 0.054 & 10.9 & 0.529 \\
 & All & 1000 & 0.444 & 0.056 & 8.3 & 0.533 \\
 & All & 2000 & 0.438 & 0.062 & 14.2 & 0.533 \\
 & Attn (all) & 100 & 0.502 & 0.005 & 7.4 & 0.553 \\
 & Attn (all) & 250 & 0.491 & 0.009 & 11.4 & 0.541 \\
 & Attn (all) & 500 & 0.483 & 0.017 & 16.2 & 0.545 \\
 & Attn (all) & 1000 & 0.470 & 0.030 & 22.6 & 0.546 \\
 & Attn (all) & 2000 & 0.460 & 0.040 & 5.8 & 0.546 \\
 & Attn L0 & 500 & 0.485 & 0.015 & 9.2 & 0.547 \\
 & Attn L1 & 100 & 0.500 & 0.002 & 11.0 & 0.576 \\
 & Attn L1 & 250 & 0.499 & 0.002 & 11.8 & 0.550 \\
 & Attn L1 & 500 & 0.495 & 0.005 & 10.3 & 0.545 \\
 & Attn L1 & 1000 & 0.490 & 0.010 & 20.4 & 0.547 \\
 & Attn L1 & 2000 & 0.487 & 0.016 & 6.4 & 0.543 \\
 & Attn L2 & 100 & 0.508 & 0.009 & 14.0 & 0.571 \\
 & Attn L2 & 250 & 0.509 & 0.009 & 13.1 & 0.561 \\
 & Attn L2 & 500 & 0.508 & 0.009 & 12.7 & 0.556 \\
 & Attn L2 & 1000 & 0.497 & 0.008 & 10.6 & 0.555 \\
 & Attn L2 & 2000 & 0.489 & 0.011 & 6.7 & 0.549 \\
 & FFN Layer 0 & 100 & 0.469 & 0.031 & 13.5 & 0.552 \\
 & FFN Layer 0 & 250 & 0.475 & 0.025 & 13.9 & 0.535 \\
 & FFN Layer 0 & 500 & 0.468 & 0.032 & 6.2 & 0.539 \\
 & FFN Layer 1 & 100 & 0.470 & 0.030 & 10.4 & 0.561 \\
 & FFN Layer 1 & 250 & 0.470 & 0.030 & 14.9 & 0.546 \\
 & FFN Layer 1 & 500 & 0.469 & 0.031 & 12.2 & 0.542 \\
 & FFN Layer 2 & 100 & 0.464 & 0.036 & 10.4 & 0.588 \\
 & FFN Layer 2 & 250 & 0.455 & 0.045 & 12.0 & 0.583 \\
 & FFN Layer 2 & 500 & 0.445 & 0.055 & 14.0 & 0.574 \\
\midrule
\multirow{12}{*}{HalfCheetah} & All & 100 & 0.349 & 0.151 & 2.3 & 0.439 \\
 & All & 250 & 0.301 & 0.199 & 2.0 & 0.439 \\
 & All & 500 & 0.224 & 0.276 & 1.6 & 0.451 \\
 & Attn (all) & 100 & 0.388 & 0.112 & 1.2 & 0.474 \\
 & Attn (all) & 250 & 0.378 & 0.122 & 1.3 & 0.480 \\
 & Attn (all) & 500 & 0.383 & 0.117 & 1.2 & 0.496 \\
 & Attn L1 & 100 & 0.439 & 0.061 & 1.2 & 0.489 \\
 & Attn L1 & 250 & 0.383 & 0.117 & 1.3 & 0.478 \\
 & Attn L1 & 500 & 0.378 & 0.122 & 1.3 & 0.481 \\
 & Attn L2 & 100 & 0.537 & 0.037 & 1.6 & 0.560 \\
 & Attn L2 & 250 & 0.497 & 0.003 & 1.6 & 0.537 \\
 & Attn L2 & 500 & 0.463 & 0.037 & 1.4 & 0.514 \\
\midrule
\multirow{12}{*}{Walker2D} & All & 100 & 0.500 & 0.013 & 7.3 & 0.445 \\
 & All & 250 & 0.512 & 0.012 & 9.1 & 0.445 \\
 & All & 500 & 0.495 & 0.015 & 6.7 & 0.446 \\
 & Attn (all) & 100 & 0.526 & 0.026 & 8.6 & 0.461 \\
 & Attn (all) & 250 & 0.505 & 0.009 & 0.2 & 0.448 \\
 & Attn (all) & 500 & 0.472 & 0.028 & 1.2 & 0.433 \\
 & Attn L1 & 100 & 0.555 & 0.055 & 10.2 & 0.477 \\
 & Attn L1 & 250 & 0.546 & 0.046 & 4.2 & 0.479 \\
 & Attn L1 & 500 & 0.538 & 0.038 & 0.6 & 0.471 \\
 & Attn L2 & 100 & 0.557 & 0.057 & 6.6 & 0.477 \\
 & Attn L2 & 250 & 0.562 & 0.062 & 0.3 & 0.473 \\
 & Attn L2 & 500 & 0.542 & 0.042 & -0.0 & 0.465 \\
\bottomrule
\end{tabular}%
}
\end{table}

%% file: figures/TABLE_A13_hopper_mlp_selective.tex
\begin{table}[htp]
\centering
\caption{Single-environment matched selective comparison in Hopper under the shared $\lambda=1.0$ setup. Each row compares the same ascent-step budget across the uniform update, the best attention-targeted edit available in the selective sweep, and the best FFN-targeted edit available in the supplementary Hopper sweep. Forget gap = $|\AUCf-0.5|$.}
\label{tab:hopper_mlp_selective}
\resizebox{1\textwidth}{!}{%
\begin{tabular}{rp{1cm}p{1.5cm}p{1.5cm}p{1.5cm}p{1.5cm}p{1.5cm}p{3.5cm}}
\toprule
Steps & Uniform Forget Gap & Uniform Utility & Best attention target & Attention Forget Gap & Attention Utility & Best FFN target & FFN Forget Gap / Utility \\
\midrule
100 & 0.020 & 6.88 & Attn L1 & 0.002 & 10.99 & FFN Layer 1 & 0.030 / 10.37 \\
250 & 0.031 & 9.93 & Attn L1 & 0.002 & 11.79 & FFN Layer 0 & 0.025 / 13.87 \\
500 & 0.054 & 10.86 & Attn L1 & 0.005 & 10.28 & FFN Layer 1 & 0.031 / 12.23 \\
\bottomrule
\end{tabular}%
}
\end{table}

%% file: figures/TABLE_A12_selective_budget.tex
\begin{table}
	\centering
	\caption{Utility-constrained selective comparison against the matched \textbf{All} baseline. A row is feasible when the D4RL normalized-score drop of the selected target relative to the uniform baseline does not exceed the stated budget. This table reports a relative budget only. Absolute retained performance must be interpreted together with \Cref{tab:oracle}.}
	\label{tab:selective_budget}
	\begin{tabular}{l|lrrr}
		\toprule
		Environment & Budget & Feasible / Total & Improved / Total & Target     \\
		\midrule
		Hopper      & 0.0    & 9/15             & 9/15             & Attn L1    \\
		      & 0.5    & 9/15             & 9/15             & Attn L1    \\
		      & 1.0    & 10/15            & 10/15            & Attn L1    \\
		      & 2.0    & 10/15            & 10/15            & Attn L1    \\
		\midrule
		HalfCheetah & 0.0    & 2/9              & 2/9              & Attn L2    \\
		 & 0.5    & 6/9              & 6/9              & Attn L2    \\
		 & 1.0    & 7/9              & 7/9              & Attn L2    \\
		 & 2.0    & 9/9              & 9/9              & Attn L2    \\
		\midrule
		Walker2D    & 0.0    & 1/9              & 0/9              & Attn (all) \\
		    & 0.5    & 2/9              & 0/9              & Attn (all) \\
		    & 1.0    & 2/9              & 0/9              & Attn (all) \\
		    & 2.0    & 2/9              & 0/9              & Attn (all) \\
		\bottomrule
	\end{tabular}
\end{table}

%% file: figures/TABLE_A14_localization_formal.tex
\begin{table}
	\centering
	\caption{Formal pre-unlearning gradient localization summary from the $\lambda=1.0$, $500$-step GA+Refit runs (3 seeds per environment). For each seed, the top target is the component with the largest ratio between gradient norms on pre-update minibatches from the forget set and the retain set, measured before the first update using two shared minibatches.}
	\label{tab:localization_formal}
	\resizebox{0.76\textwidth}{!}{%
		\begin{tabular}{l|ccrrl}
			\toprule
			Environment & Dominant Top Target & Count & Mean Top Ratio & Mean Ratio Span & Unique Top Targets   \\
			\midrule
			HalfCheetah & FFN Layer 2         & 3/3   & 1.307          & 0.373           & FFN Layer 2          \\
			Hopper      & FFN Layer 0         & 2/3   & 2.071          & 0.995           & All, FFN Layer 0     \\
			Walker2D    & FFN Layer 1         & 2/3   & 1.182          & 0.254           & Attn L1, FFN Layer 1 \\
			\bottomrule
		\end{tabular}%
	}
\end{table}

%% file: figures/TABLE_A4_matching_power.tex
\begin{table}[htp]
\centering
\caption{Matching diagnostics and precision summary for the selected high-utility comparative settings. Matching removes the largest return and length imbalances present before pairing, but some initial-state discrepancies remain. The absolute standardized mean difference (SMD) columns report the remaining imbalance after matching for return, trajectory length, the state$_0$ norm, and the maximum coordinate-level state$_0$ discrepancy. The final four columns summarize the effective sample size, the hierarchical 95\% interval for the base DT forget-set AUC, the mean paired-bootstrap interval width across seeds, and the approximate detectable AUC gap under a balanced two-sided test with $\alpha=0.05$ and power $0.80$. These diagnostics characterize pair-level and seed-aware uncertainty and should be interpreted separately from the seed-level intervals reported in the main tables.}
\label{tab:matching_power}
\resizebox{\textwidth}{!}{%
\begin{tabular}{l|cp{2.3cm}p{2.5cm}p{2.5cm}p{2.5cm}p{1.3cm}cp{2cm}p{2cm}}
\toprule
Environment & Match rate & Return abs SMD            & Length abs SMD            & State$_0$ norm abs SMD    & State$_0$ coord max abs SMD & Mean pairs & Hierarchical 95\% CI & Mean paired CI width & Detectable Forget Gap \\
\midrule
HalfCheetah (ME) & 140/140 & 0.078 $\rightarrow$ 0.002 & 0.000 $\rightarrow$ 0.000 & 0.037 $\rightarrow$ 0.020 & 0.181 $\rightarrow$ 0.207 & 140 & [0.609, 0.672] & 0.110 & 0.097 \\
Walker2D (ME) & 153/153 & 0.041 $\rightarrow$ 0.004 & 0.054 $\rightarrow$ 0.006 & 0.105 $\rightarrow$ 0.045 & 0.179 $\rightarrow$ 0.237 & 153 & [0.499, 0.534] & 0.060 & 0.093 \\
Walker2D (M) & 83/83 & 0.346 $\rightarrow$ 0.003 & 0.334 $\rightarrow$ 0.005 & 0.242 $\rightarrow$ 0.082 & 0.236 $\rightarrow$ 0.365 & 83 & [0.591, 0.651] & 0.104 & 0.126 \\
\bottomrule
\end{tabular}%
}
\end{table}

%% file: figures/TABLE_A15_matching_variants.tex
\begin{table}[htp]
\centering
\caption{Matching-robustness comparison for the canonical medium-replay benchmark methods in Hopper and Walker2D. Privacy is summarized by the mean forget gap across the three canonical benchmark seeds for each benchmark block. The stronger variant augments the matching feature space with reward, action, and trajectory-dynamics summaries while preserving one-to-one pairing. The unmatched baseline removes the matching control entirely by pairing the forget trajectories directly with the first test trajectories.}
\label{tab:matching_variant}
\resizebox{\textwidth}{!}{%
\begin{tabular}{l|lccccc}
\toprule
Environment & Method & Seeds & Basic Forget Gap & Stronger Forget Gap & Unmatched Forget Gap & Stronger$-$Basic $\Delta$ \\
\midrule
\multirow{4}{*}{Hopper} & Base DT (B1) & 3 & 0.023 & 0.036 & 0.005 & 0.014 \\
 & Retrain Ref. (B2) & 3 & 0.018 & 0.032 & 0.003 & 0.014 \\
 & Naive FT (B3) & 3 & 0.016 & 0.024 & 0.010 & 0.008 \\
 & GA+Refit (B4) & 3 & 0.052 & 0.030 & 0.070 & -0.022 \\
\midrule
\multirow{4}{*}{Walker2D} & Base DT (B1) & 3 & 0.125 & 0.065 & 0.124 & -0.061 \\
 & Retrain Ref. (B2) & 3 & 0.020 & 0.008 & 0.059 & -0.012 \\
 & Naive FT (B3) & 3 & 0.028 & 0.005 & 0.066 & -0.022 \\
 & GA+Refit (B4) & 3 & 0.089 & 0.054 & 0.121 & -0.035 \\
\bottomrule
\end{tabular}%
}
\end{table}

%% file: figures/TABLE_A9_tost_sensitivity.tex
\begin{table}
\centering
\caption{TOST equivalence test pass rates ($p_{\text{TOST}} < 0.05$) under varying margins $\varepsilon$. The test evaluates whether the forget-set AUC is equivalent to random guessing ($0.5$) within the stated margin. Highlighted rows denote the component-level configurations used by this TOST analysis for each environment (Attn L1, Attn L2, All Attn). These rows are three-seed summaries built from the cross-validated target family reported in \Cref{tab:layer_cv}: for each environment, the target is fixed to the cross-validated choice and the ascent-step is then selected from the `best\_matches\_cv=True` candidates in the component-level stability analysis. Pass patterns at the strictest margin ($\varepsilon = 0.05$) remain environment and method dependent.}
\label{tab:tost_sensitivity}
\resizebox{\textwidth}{!}{%
\begin{tabular}{l|lrcccc}
\toprule
Env & Method & Forget AUC & $\varepsilon=0.05$ & $\varepsilon=0.075$ & $\varepsilon=0.10$ & $\varepsilon=0.15$ \\
\midrule
\multirow{4}{*}{Hopper} & Retrain Ref. & 0.518 & $\checkmark$ & $\checkmark$ & $\checkmark$ & $\checkmark$ \\
 & Naive FT & 0.516 & $\checkmark$ & $\checkmark$ & $\checkmark$ & $\checkmark$ \\
 & GA+Refit & 0.510 & $\times$ & $\times$ & $\checkmark$ & $\checkmark$ \\
 & \textbf{Selective (Attn L1)} & \textbf{0.499} & $\checkmark$ & $\checkmark$ & $\checkmark$ & $\checkmark$ \\
\midrule
\multirow{4}{*}{HalfCheetah} & Retrain Ref. & 0.498 & $\checkmark$ & $\checkmark$ & $\checkmark$ & $\checkmark$ \\
 & Naive FT & 0.500 & $\checkmark$ & $\checkmark$ & $\checkmark$ & $\checkmark$ \\
 & GA+Refit & 0.784 & $\times$ & $\times$ & $\times$ & $\times$ \\
 & \textbf{Selective (Attn L2)} & \textbf{0.497} & $\checkmark$ & $\checkmark$ & $\checkmark$ & $\checkmark$ \\
\midrule
\multirow{4}{*}{Walker2D} & Retrain Ref. & 0.520 & $\checkmark$ & $\checkmark$ & $\checkmark$ & $\checkmark$ \\
 & Naive FT & 0.528 & $\checkmark$ & $\checkmark$ & $\checkmark$ & $\checkmark$ \\
 & GA+Refit & 0.589 & $\times$ & $\times$ & $\checkmark$ & $\checkmark$ \\
 & \textbf{Selective (All Attn)} & \textbf{0.505} & $\checkmark$ & $\checkmark$ & $\checkmark$ & $\checkmark$ \\
\bottomrule
\end{tabular}%
}
\end{table}

%% file: figures/TABLE_A10_layer_cv.tex
\begin{table}
\centering
\caption{Leave-one-seed-out cross-validation for layer selection. Hopper and HalfCheetah select the same attention layer in every split, whereas Walker2D does not show a stable or consistent advantage over the uniform baseline across held-out seeds. The CV gap is reported per held-out split and therefore is not numerically identical to the budget-filtered aggregate in \Cref{tab:oracle}.}
\label{tab:layer_cv}
\begin{tabular}{l|lllrr}
\toprule
Env & Held-out & CV Selected & Oracle & CV Forget Gap & Uniform Forget Gap \\
\midrule
\multirow{3}{*}{Hopper} & seed 0 & Attn L1 & Attn L1 & 0.003 & 0.030 \\
 & seed 1 & Attn L1 & Attn L1 & 0.003 & 0.038 \\
 & seed 2 & Attn L1 & Attn L1 & 0.003 & 0.037 \\
\midrule
\multirow{3}{*}{HalfCheetah} & seed 0 & Attn L2 & Attn L2 & 0.026 & 0.187 \\
 & seed 1 & Attn L2 & Attn L2 & 0.027 & 0.218 \\
 & seed 2 & Attn L2 & Attn L2 & 0.026 & 0.221 \\
\midrule
\multirow{3}{*}{Walker2D} & seed 0 & Attn (all) & Attn (all) & 0.028 & 0.012 \\
 & seed 1 & Attn (all) & Attn (all) & 0.015 & 0.021 \\
 & seed 2 & Attn (all) & Attn (all) & 0.020 & 0.006 \\
\bottomrule
\end{tabular}
\end{table}

%% file: figures/TABLE_A5_action_error.tex
\begin{table}[htp]
\centering
\caption{Action-error membership attack on the selected high-utility comparative settings aggregated across seeds. The score is the mean squared error between predicted and observed actions. Values closer to $0.5$ indicate weaker membership leakage. The reported $p$-value is a two-sided permutation test for the null hypothesis that the attack scores of forget trajectories and matched non-members are exchangeable.}
\label{tab:action_error_attack}
\resizebox{\textwidth}{!}{%
\begin{tabular}{l|lrrrrr}
\toprule
Environment                  & Method        & Utility $\uparrow$ & Forget AUC   & Forget Gap $\downarrow$ & 95\% CI width & $p$-value \\
\midrule
\multirow{5}{*}{HalfCheetah (ME)} & Base DT & 64.13 & 0.657 & 0.157 & 0.130 & 0.0001 \\
 & Retrain Ref. & 59.82 & 0.563 & 0.063 & 0.137 & 0.0754 \\
 & Naive FT & 60.63 & 0.561 & 0.061 & 0.137 & 0.0812 \\
 & GA+Refit & 4.70 & 0.589 & 0.089 & 0.134 & 0.0135 \\
 & TrajDeleter & 58.74 & 0.638 & 0.138 & 0.132 & 0.0002 \\
\midrule
\multirow{5}{*}{Walker2D (ME)} & Base DT & 93.11 & 0.512 & 0.012 & 0.131 & 0.7268 \\
 & Retrain Ref. & 87.11 & 0.481 & 0.019 & 0.131 & 0.5688 \\
 & Naive FT & 85.46 & 0.479 & 0.021 & 0.131 & 0.5357 \\
 & GA+Refit & 0.00 & 0.515 & 0.015 & 0.130 & 0.6483 \\
 & TrajDeleter & 92.73 & 0.506 & 0.006 & 0.131 & 0.8627 \\
\midrule
\multirow{5}{*}{Walker2D (M)} & Base DT & 71.92 & 0.614 & 0.114 & 0.170 & 0.0117 \\
 & Retrain Ref. & 67.62 & 0.546 & 0.046 & 0.175 & 0.3098 \\
 & Naive FT & 77.63 & 0.546 & 0.046 & 0.176 & 0.3121 \\
 & GA+Refit & 13.50 & 0.604 & 0.104 & 0.172 & 0.0208 \\
 & TrajDeleter & 67.90 & 0.601 & 0.101 & 0.172 & 0.0254 \\
\bottomrule
\end{tabular}%
}
\end{table}

%% file: figures/TABLE_A6_representation_shadow.tex
\begin{table}[htp]
\centering
\caption{Supplementary attack-family diagnostics on the selected high-utility comparative settings with expanded seed coverage. The upper panel reports the representation attack averaged across seeds using final-layer state-token features. Values closer to $0.5$ indicate weaker membership leakage, whereas values below $0.5$ indicate an inverted ranking rather than stronger deletion. The reported $p$-value is a two-sided permutation test for the null hypothesis that the representation-based scores of forget trajectories and matched non-members are exchangeable. The lower panel reports a target-seed-rotated held-out shadow-model attack with eight queried steps per trajectory, averaged over target seeds. Because the retraining reference itself remains meaningfully above $0.5$ under this restricted attack in multiple environments, the lower panel is reported together with a retrain-relative excess gap, defined as the residual forget gap after subtracting the retraining-reference gap within the same environment. The shadow-query panel should therefore be interpreted as an exploratory diagnostic of attack-family sensitivity rather than as a standalone forgetting-validity criterion.}
\label{tab:representation_shadow_attack}
\resizebox{\textwidth}{!}{%
\begin{tabular}{l|l|lrrrr}
\toprule
Attack                           & Environment                  & Method        & Utility $\uparrow$ & Forget AUC   & Forget Gap $\downarrow$ & $p$-value \\
\midrule
\multirow{15}{*}{Representation} & \multirow{5}{*}{HalfCheetah (ME)} & Base DT & 64.13 & 0.450 & 0.050 & 0.2238 \\
 &  & Retrain Ref. & 59.82 & 0.450 & 0.050 & 0.2574 \\
 &  & Naive FT & 60.63 & 0.466 & 0.034 & 0.3796 \\
 &  & GA+Refit & 4.70 & 0.524 & 0.024 & 0.5495 \\
 &  & TrajDeleter & 58.74 & 0.458 & 0.042 & 0.2597 \\
\cmidrule(lr){2-7}
 & \multirow{5}{*}{Walker2D (ME)} & Base DT & 93.11 & 0.557 & 0.057 & 0.1566 \\
 &  & Retrain Ref. & 87.11 & 0.542 & 0.042 & 0.2858 \\
 &  & Naive FT & 85.46 & 0.542 & 0.042 & 0.3160 \\
 &  & GA+Refit & 0.00 & 0.500 & 0.004 & 0.9063 \\
 &  & TrajDeleter & 92.73 & 0.558 & 0.058 & 0.1623 \\
\cmidrule(lr){2-7}
 & \multirow{5}{*}{Walker2D (M)} & Base DT & 71.92 & 0.463 & 0.037 & 0.4214 \\
 &  & Retrain Ref. & 67.62 & 0.471 & 0.029 & 0.5418 \\
 &  & Naive FT & 77.63 & 0.477 & 0.023 & 0.6351 \\
 &  & GA+Refit & 13.50 & 0.406 & 0.094 & 0.0831 \\
 &  & TrajDeleter & 67.90 & 0.467 & 0.033 & 0.4686 \\
\midrule
\multicolumn{7}{l}{\textit{Shadow-query panel: the last column reports the retrain-relative excess gap rather than the permutation $p$-value.}} \\
\cmidrule(lr){1-7}
Attack                           & Environment                  & Method        & Utility $\uparrow$ & Forget AUC   & Forget Gap $\downarrow$ & Excess over retrain $\downarrow$ \\
\midrule
\multirow{15}{*}{Shadow query} & \multirow{5}{*}{HalfCheetah (ME)} & Base DT & 64.13 & 0.539 & 0.039 & -0.001 \\
 &  & Retrain Ref. & 59.82 & 0.541 & 0.041 & 0.000 \\
 &  & Naive FT & 60.63 & 0.555 & 0.055 & 0.015 \\
 &  & GA+Refit & 4.70 & 0.559 & 0.059 & 0.018 \\
 &  & TrajDeleter & 58.74 & 0.538 & 0.038 & -0.002 \\
\cmidrule(lr){2-7}
 & \multirow{5}{*}{Walker2D (ME)} & Base DT & 93.11 & 0.586 & 0.086 & 0.004 \\
 &  & Retrain Ref. & 87.11 & 0.582 & 0.082 & 0.000 \\
 &  & Naive FT & 85.46 & 0.572 & 0.072 & -0.010 \\
 &  & GA+Refit & 0.00 & 0.566 & 0.066 & -0.016 \\
 &  & TrajDeleter & 92.73 & 0.589 & 0.089 & 0.007 \\
\cmidrule(lr){2-7}
 & \multirow{5}{*}{Walker2D (M)} & Base DT & 71.92 & 0.662 & 0.162 & 0.006 \\
 &  & Retrain Ref. & 67.62 & 0.656 & 0.156 & 0.000 \\
 &  & Naive FT & 77.63 & 0.656 & 0.156 & -0.001 \\
 &  & GA+Refit & 13.50 & 0.577 & 0.077 & -0.079 \\
 &  & TrajDeleter & 67.90 & 0.649 & 0.149 & -0.007 \\
\bottomrule
\end{tabular}%
}
\end{table}

%% file: figures/TABLE_A11_fisher_baseline.tex
\begin{table}[htp]
\centering
\caption{Diagonal-Fisher unlearning baseline on the selected high-utility comparative settings, aggregated across seeds $0,1,2$ with $damping=10^{-3}$, $scale=1.0$, and head refitting enabled. Across all three settings, Fisher approaches random-guessing privacy only by collapsing utility, so it is retained as a failure-mode baseline rather than a competitive alternative.}
\label{tab:fisher_baseline}
\begin{tabular}{l|rrrrr}
\toprule
Environment & Seeds & Utility $\uparrow$ & Forget AUC & Forget Gap $\downarrow$ & Retrain Valid \\
\midrule
HalfCheetah (ME) & 3 & 1.71 & 0.489 & 0.011 & \cmark \\
Walker2D (ME) & 3 & 2.61 & 0.468 & 0.032 & \cmark \\
Walker2D (M) & 3 & 0.12 & 0.535 & 0.035 & \cmark \\
\bottomrule
\end{tabular}
\end{table}